\documentclass[twoside,twocolumn,9pt]{article}
\usepackage{extsizes}
\usepackage[super,sort&compress,comma]{natbib} 
\usepackage[version=3]{mhchem}
\usepackage[left=1.5cm, right=1.5cm, top=1.785cm, bottom=2.0cm]{geometry}
\usepackage{balance}
\usepackage{mathptmx}
\usepackage{amssymb}
\usepackage{sectsty}
\usepackage{graphicx} 
\usepackage{lastpage}
\usepackage[format=plain,justification=justified,singlelinecheck=false,font={stretch=1.125,small,sf},labelfont=bf,labelsep=space]{caption}
\usepackage{float}
\usepackage{fancyhdr}
\usepackage{fnpos}
\usepackage[english]{babel}
\addto{\captionsenglish}{%
  
}
\usepackage{array}
\usepackage{droidsans}
\usepackage{charter}
\usepackage[T1]{fontenc}
\usepackage[usenames,dvipsnames]{xcolor}
\usepackage{setspace}
\usepackage[compact]{titlesec}
\usepackage{hyperref}
\definecolor{cream}{RGB}{222,217,201}

\begin{document}

\pagestyle{fancy}
\thispagestyle{plain}
\fancypagestyle{plain}{
%%%HEADER%%%
\renewcommand{\headrulewidth}{0pt}
}
%%%END OF HEADER%%%

%%%PAGE SETUP - Please do not change any commands within this section%%%
\makeFNbottom
\makeatletter
\renewcommand\LARGE{\@setfontsize\LARGE{15pt}{17}}
\renewcommand\Large{\@setfontsize\Large{12pt}{14}}
\renewcommand\large{\@setfontsize\large{10pt}{12}}
\renewcommand\footnotesize{\@setfontsize\footnotesize{7pt}{10}}
\makeatother

\renewcommand{\thefootnote}{\fnsymbol{footnote}}
\renewcommand\footnoterule{\vspace*{1pt}% 
\color{cream}\hrule width 3.5in height 0.4pt \color{black}\vspace*{5pt}} 
\setcounter{secnumdepth}{5}

\makeatletter 
\renewcommand\@biblabel[1]{#1}            
\renewcommand\@makefntext[1]% 
{\noindent\makebox[0pt][r]{\@thefnmark\,}#1}
\makeatother 
\renewcommand{\figurename}{\small{Fig.}~}
\sectionfont{\sffamily\Large}
\subsectionfont{\normalsize}
\subsubsectionfont{\bf}
\setstretch{1.125} %In particular, please do not alter this line.
\setlength{\skip\footins}{0.8cm}
\setlength{\footnotesep}{0.25cm}
\setlength{\jot}{10pt}
\titlespacing*{\section}{0pt}{4pt}{4pt}
\titlespacing*{\subsection}{0pt}{15pt}{1pt}
%%%END OF PAGE SETUP%%%

%%%FOOTER%%%
\fancyfoot{}
\fancyfoot[LO,RE]{\vspace{-7.1pt}\includegraphics[height=9pt]{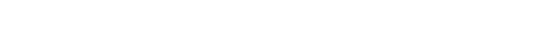}}
\fancyfoot[CO]{\vspace{-7.1pt}\hspace{13.2cm}\includegraphics{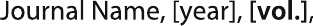}}
\fancyfoot[CE]{\vspace{-7.2pt}\hspace{-14.2cm}\includegraphics{head_foot/RF}}
\fancyfoot[RO]{\footnotesize{\sffamily{1--\pageref{LastPage} ~\textbar  \hspace{2pt}\thepage}}}
\fancyfoot[LE]{\footnotesize{\sffamily{\thepage~\textbar\hspace{3.45cm} 1--\pageref{LastPage}}}}
\fancyhead{}
\renewcommand{\headrulewidth}{0pt} 
\renewcommand{\footrulewidth}{0pt}
\setlength{\arrayrulewidth}{1pt}
\setlength{\columnsep}{6.5mm}
\setlength\bibsep{1pt}
%%%END OF FOOTER%%%

%%%FIGURE SETUP - please do not change any commands within this section%%%
\makeatletter 
\newlength{\figrulesep} 
\setlength{\figrulesep}{0.5\textfloatsep} 

\newcommand{\topfigrule}{\vspace*{-1pt}% 
\noindent{\color{cream}\rule[-\figrulesep]{\columnwidth}{1.5pt}} }

\newcommand{\botfigrule}{\vspace*{-2pt}% 
\noindent{\color{cream}\rule[\figrulesep]{\columnwidth}{1.5pt}} }

\newcommand{\dblfigrule}{\vspace*{-1pt}% 
\noindent{\color{cream}\rule[-\figrulesep]{\textwidth}{1.5pt}} }

\makeatother
%%%END OF FIGURE SETUP%%%

%%%TITLE, AUTHORS AND ABSTRACT%%%
%%% !!!! uncommmen this when submitting!!!
\twocolumn[
  \begin{@twocolumnfalse}
{%\includegraphics[height=30pt]{head_foot/journal_name}\hfill\raisebox{0pt}[0pt][0pt]{\includegraphics[height=55pt]{head_foot/RSC_LOGO_CMYK}}\\[1ex]
}\par
\vspace{1em}
\sffamily
\begin{tabular}{m{4.5cm} p{13.5cm} }

\includegraphics{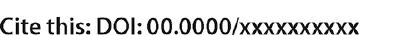} & \noindent\LARGE{\textbf{Shape is (almost) all!: Persistent homology features (PHFs) are an information rich input for efficient molecular machine learning.$^\dag$}} \\%Article title goes here instead of the text "This is the title"
\vspace{0.3cm} & \vspace{0.3cm} \\

 & \noindent\large{Ella M. Gale$^{\ast}$\textit{$^{a}$} } \\%Author names go here instead of "Full name", etc.

\includegraphics{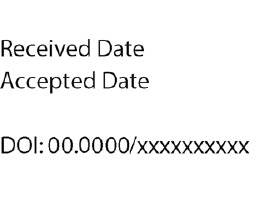} & \noindent\normalsize{3-D shape is important to chemistry, but how important? Machine learning works best when the inputs are simple and match the problem well. Chemistry datasets tend to be very small compared to those generally used in machine learning so we need to get the most from each datapoint. Persistent homology measures the topological shape properties of point clouds at different scales and is used in topological data analysis. Here we investigate what persistent homology captures about molecular structure and create persistent homology features (PHFs) that encode a molecule's shape whilst losing most of the symbolic detail like atom labels, valence, charge, bonds etc. We demonstrate the usefulness of PHFs on a series of chemical datasets: QM7,  lipophilicity, Delaney and Tox21. PHFs work as well as the best benchmarks. PHFs are very information dense and much smaller than other encoding methods yet found, meaning ML algorithms are much more energy efficient. PHFs success despite losing a large amount of chemical detail highlights how much of chemistry can be simplified to topological shape.} \\%The abstrast goes here instead of the text "The abstract should be..."

\end{tabular}

 \end{@twocolumnfalse} \vspace{0.6cm}

  ]
%%%END OF TITLE, AUTHORS AND ABSTRACT%%%

%%%FONT SETUP - please do not change any commands within this section
\renewcommand*\rmdefault{bch}\normalfont\upshape
\rmfamily
\section*{}
\vspace{-1cm}

%%%FOOTNOTES%%%

\footnotetext{\textit{$^{a}$~School of Chemistry, University of Bristol, Cantock's Close, Bristol, UK.  E-mail: ella.gale@bristol.ac.uk}}
%\footnotetext{\textit{$^{b}$~Address, Address, Town, Country. }}

%Please use \dag to cite the ESI in the main text of the article.
%If you article does not have ESI please remove the the \dag symbol from the title and the footnotetext below.
\footnotetext{\dag~Electronic Supplementary Information (ESI) available: [details of any supplementary information available should be included here]. See DOI: 00.0000/00000000.}

\section{Introduction}

What is a bond? The existence of atoms and bonds was hypothesized long ago, atoms have been found to have a physical reality but bonds have not,\cite{brown2002topology, coulson1955contributions, weisberg2008challenges} and functional groups derived from them are not the autonomous objects often assumed.\cite{mezey2012molecular} Bonds are a useful short-cut for chemists and the simple rules formulated to describe them often have counterexamples. So what then is a molecule? And which of the many computer-accessible descriptions of a molecule is the best for a specific problem? Finally, what can the success or failure of a specific description tell us about the chemistry of a specific problem?

Machine learning is best described as discovering, manipulating and using patterns in high-dimensional data. Molecular machine learning (MML) is the application of machine learning approaches to chemical problems. The choice of featurisation (molecular description for input into a ML algorithm) is crucial and not obvious. There are many featurisations available, fingerprints based on properties of the molecule\cite{gluck1965chemical,rogers2010extended, lowis1998hqsar, durant2002reoptimization}, physicochemical properties,\cite{murray2016application,moseley2014ligand} DFT\cite{durand2019computational} etc. Chemistry is inherently at least a 3-D problem. The shape of molecules is critical to their function, for example a drug's action, chirality, optical, electronic structure, reactivity, available reaction pathways, frustration and many more chemical concepts are all inherently based on shape. A key motivation for my work is that an ML algorithm will solve problems in a more efficient and stable manner if we give them the correct information in the most usable manner, i.e. if we make the problem as easy as possible for the algorithm. For problems involving 3-D properties, this requires a featurisation that is either 3-D or encodes 3-D properties. How to encode this shape in a machine readable format for use in machine learning is not yet a solved problem. 

%Any chemist knows that the 3-D structure of a molecule is critical to its properties (symmetry, chirality etc) and uses (for example, in binding to proteins), and, as drug design starts to move away from molecules easily describable as 2-D (flat) to 3-D, the requirement for good 3-D descriptions is more pressing and is considered an open problem. MML already has several useful 3-D featurisations, with their own advantages and drawbacks. 

%3-D Cartesian coordinates or voxels do not work very well as they are a sparse input (which neural networks find hard to learn from).

Early artificial intelligence in general, and chemistry in specific, used symbols as inputs to create symbolic AI. Recently, it has been found that more than 10,000 rules were required to map less than half of a large chemistry reaction dataset,\cite{schwaller2021mapping10} which is too many for human experts to handcode. Recent huge successes in the fields of non-symbolic neural networks (see for example~\cite{AlexNet, 65, 47}  and other ML techniques suggest a different approach, that we can do away with symbols for some tasks.  It is interesting to ask how far this iconoclastry can be taken in chemistry, is it possible to do any useful machine learning with no human-centric symbols whatsoever? What happens if we remove atom labels, bonds, charges, valence and all other symbols and simply reduce molecules to collections of points in space? Is there anything useful that can be learned about such an approach?

In this paper we will apply a novel method of describing a molecule based on persistent homology to molecular machine learning problems. This novel method encodes the molecule's 3-D structure and shape in a manner unfamiliar to the chemist, and does not directly encode atom or bond properties. Where this approach succeeds or fails can inform our chemical understanding of a problem. This featurisation can also be usefully combined with other featurisations to give state of the art results in a highly efficient manner. 

As the the field of persistent homology is very new, not well known in chemistry and we believe, has the possibility of being very useful to the chemical sciences, this paper will include a thorough review of the literature and explanation of the application of these techniques to aid chemists in adding these methods to their tool-kits. Section 1.1 will outline where topology has already been used in chemistry to allow us to differentiate this topological method from others, section~\ref{sec:methodology} will outline what persistence homology is and section ~\ref{sec:benzene} demonstrates how it can be applied using chemical examples, and the full mathematical details are given in the appendix. Section~\ref{ssec:chem_prob} outlines the chemical problems investigated, section \ref{sec:featurisations} outlines what featurisations are available and reviews ML models. The results of these experiments are in section~\ref{sec:results}. Example code is in appendix~\ref{sec:code_example}. Full mathematical details are in appendix~\ref{sec:maths}. 

\subsection{Featurisations for molecular machine learning\label{sec:featurisations}}

\subsubsection{Featusization methods}

%[[this probably wants to go in icospherer]]

To input a molecule in a machine learning algorithm we must first find a way to encode the features of that molecule that are relevant to the problem (this is the process of featurisation). There is no one featurisation that is better than the others, it depends on the type of problem and the featurisation chosen directs the choice of model trained, see table~\ref{tab:featurisations}.

Despite being designed for inputting into a computer, string based encodings like SMILES, SMIRKS, SELFIES are not great for machine learning, although there has been progress on training ML models used in language processing\cite{schwaller2020predicting}, however, the output molecules tended to be `weird' or `unchemical'. 
% NTS I don't think that reference is correct, find the proper one
This has been explained by suggesting that although chemical formulae look like a language that follows a grammar this may not actually be true.\cite{balllang, gordinlang} My view is that the 3-D nature of molecules is a crucial feature that must be inputted and without some encoding of the 3-D structure the algorithm cannot learn about bond angles and torsions which is required for predicting valid chemical structures.

%[[!!! the following sections need to be here but need to be rewritten]]
%\paragraph{1-D methods} For some tasks, for example solvent selection, the structure of the molecule is not directly relevant, and a set of physicochemical properties (like boiling point) and solvent [[properties]] (like [[forogot]]) will work fine.[[cite your and murrys work]]. Molecular fingerprints are a 1-D method as the input is a vector of numbers, either a binary series of presence/absence of features (like...[[]]) or a hash of considered properties (like...[[]]). 

%Note that, all the methods discussed thus far produce a featurisation based on the local environment and local features. The power of TDAF (and topology in general) is that it uses local environemental information to describe aspects of the global properties (Specifically, the global topological invariants), which is in some way a measure of the gross shape of molecules. 

%\paragraph{2-D methods}

%graphs

%\paragraph{3-D methods}

\begin{table*}[htp]
\small
  \caption{\ Common and novel featurisations used in molecular machine learning. NN is a feed-forward NN, RF is random forest, KRR is kernel ridge regression. *This method is based on geometric machine learning and is currently under development within our group.}
  \label{tab:featurisations}
  \begin{tabular*}{\textwidth}{@{\extracolsep{\fill}}llllll}
    \hline
    Type & Dimensionality  & Dimensionality & Mathematical & Examples & Example\\
     & of input & of featurisation & basis & & models\\
    \hline
    Chemical formula & 1-D string & 1-D & N/A & SMILES, SMARTS, SMIRKS & MTR or textNN\\
    Fingerprints & 1-D string & 1-D & N/A & ECFP, MACCS & NN, RF, KRR\\
    physicochemical properties & N/A & 1-D & N/A & Rdkit & Multitask Regression\\
    Coulomb matrix & 3-D & 2-D & Graph topology & Coulomb matrix & DTNN \\
    Eigenvalues of Coulomb matrix & 3-D & 1-D & Graph topology & CM-eig & Multitask Regression\\
    Graph & 2-D & 2-D & Graph toplogy & Graph convolutions & Graph convolutional\\
    Graph & 2-D & 2-D & Graph toplogy & Weave convolutions & Weave neural network\\
    Grid & 3-D & 1-D & Geometry  & Grid &  MTR\\
   % Symmetry functions & 3-D? & symmetry & ?\\
    Icospherical/spherical$^{*}$ & 3-D & 3-D & Geometry & Icospherical maps & Spherical NN\\
  %  (forthcoming work) & 3-D & 3-D & projection & Icospherical maps &Spherical NN\\
   PHF (this work) & 3-D & 1-D & Persistent homology & PHF & Multitask Regression \\ 
    \hline
  \end{tabular*}
\end{table*}

\subsection{Topology in chemistry\label{sec:topol_review}}

\begin{figure}[htp]
 \centering
 \begin{tabular}{p{3.3cm}p{3.3cm}}
 a. A knot &b. A torus\\
     \includegraphics[width=3.2cm]{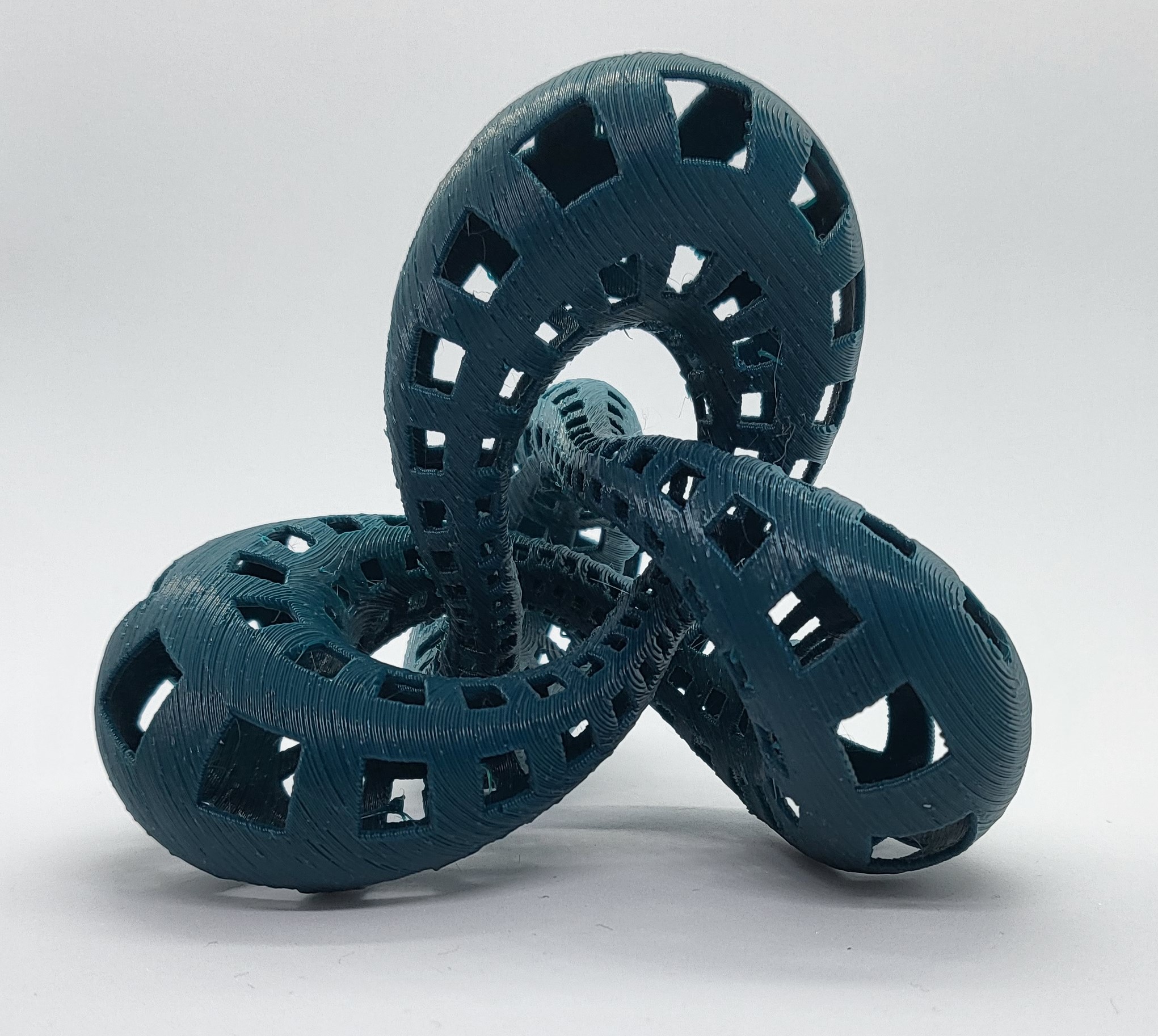} & 
     \includegraphics[width=3.2cm]{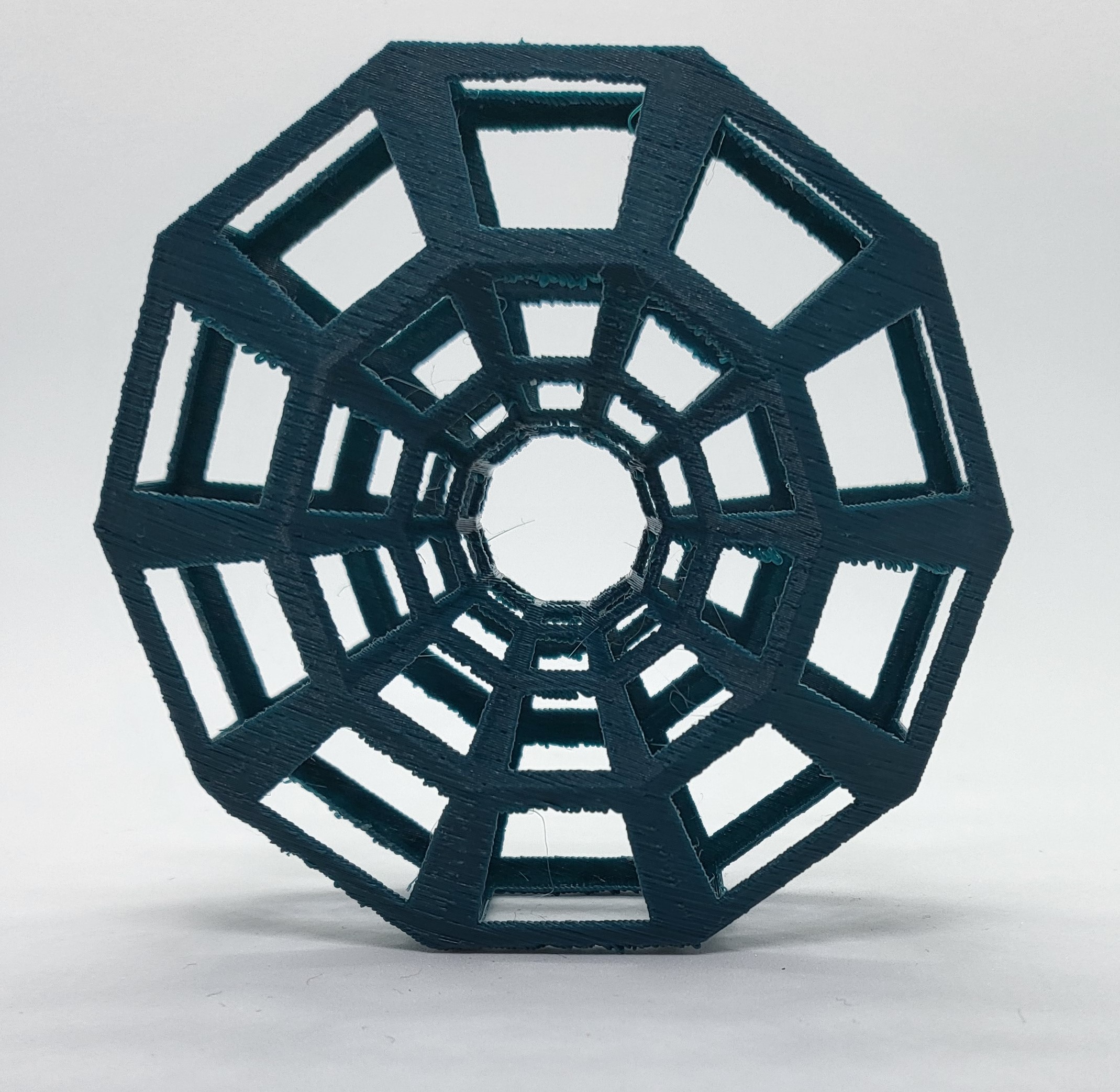}\\
    
    \includegraphics[width=3.2cm]{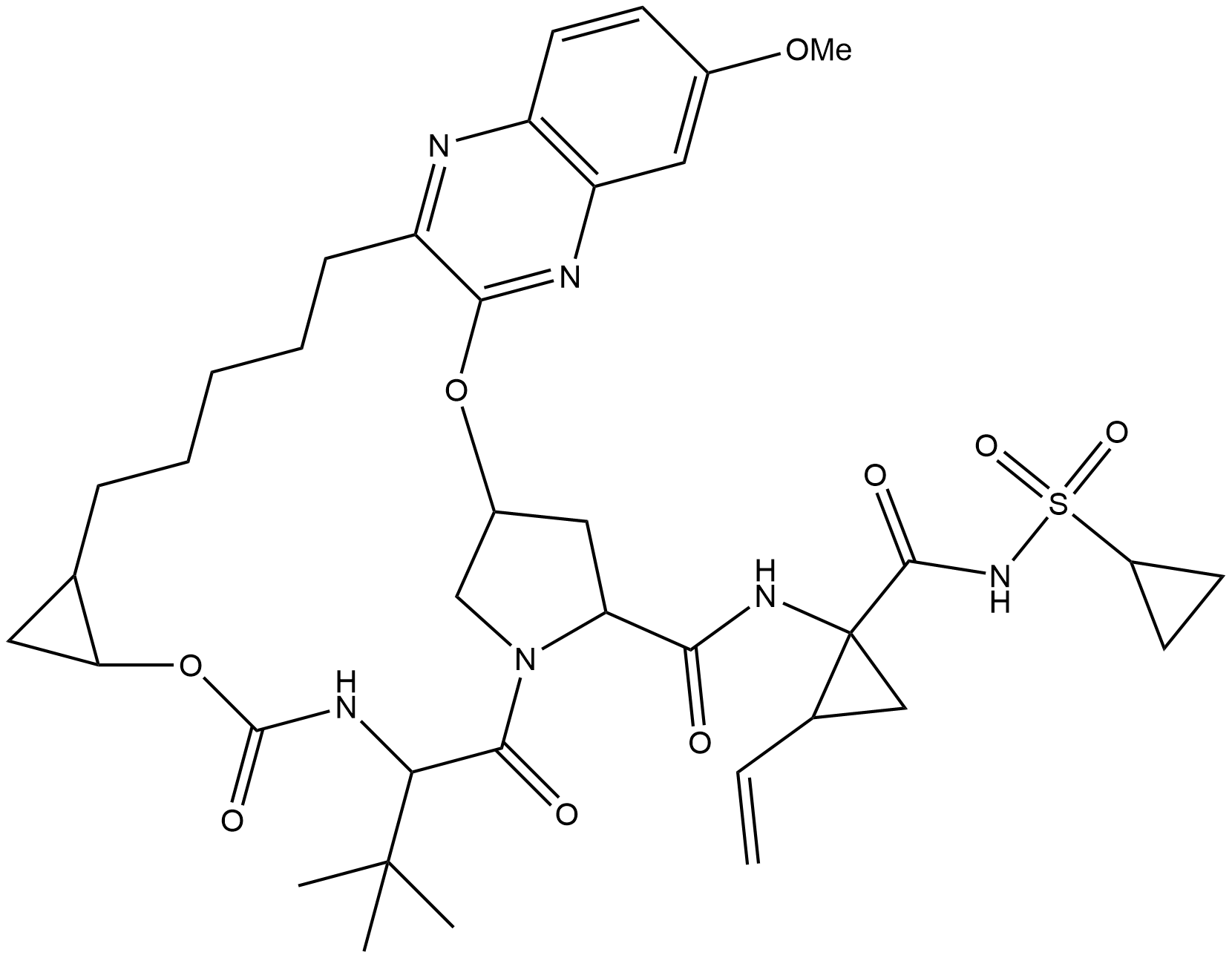} &
    \includegraphics[width=3.2cm]{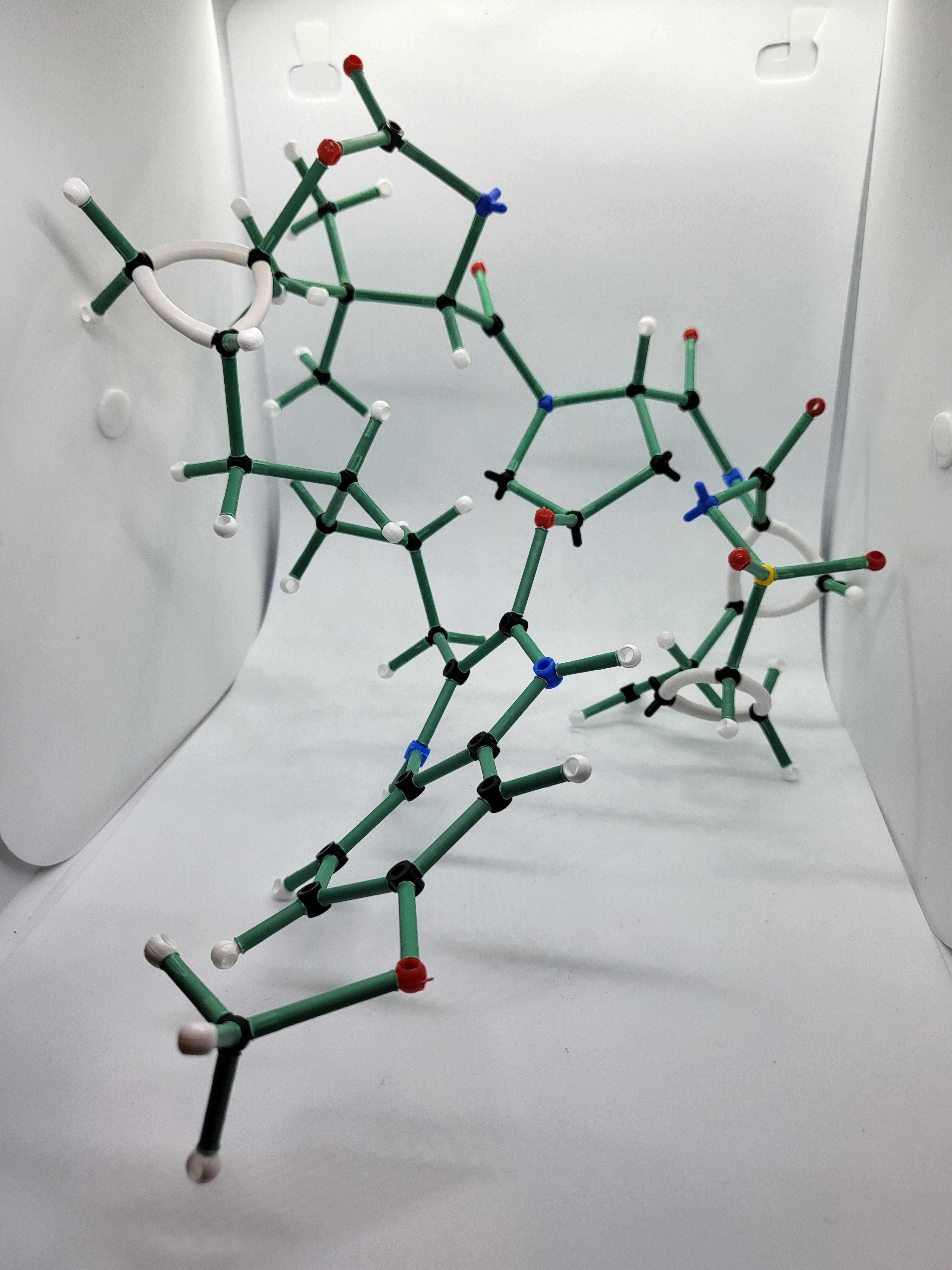}\\
    c. A graph & d. A 3-D molecular structure\\
 \end{tabular}
 \caption{Objects dealt with in topology: top left: knots (knot topology); top right: torus (homology); bottom left: graphs (graph topology); bottom right: 3-D molecular structure. Note that molecular graphs are 2-D representations and only include the connections between molecules, not stereochemistry and as such miss out the 3-D shape of the molecule. Structures are grazoprevir, a macrocyclic drug used to treat hepatitis C and is deliberately drawn without the stereochemistry as that is not included in the molecular graph and thus not available for topological analysis using graph topology.} 
 %3-D printing stls come from [[cite book]]. [[redraw the structure without stereochemisty]].}
 \label{fig:topol_example}
\end{figure}

Topology is a branch of mathematics primarily concerned with shape. Specifically, those properties which are invariant to continuous deformations (e.g. deformations like pulling, twisting, deforming and not actions like cutting, threading etc.) Several ideas from topology have already found their way into the chemical and biological literature (although the term as chemists use it is generally a lot broader than when it is used by mathematicians\cite{francl2009stretching}). To make this contribution clear, we will briefly review topological ideas and their influence on chemistry. To differentiate the separate sub-fields from one another, the area of topology will be included when topology is referred to. 

\subsubsection{Polyhedral topology} The ancient Greeks recognised five polyhedra, all of which are of great relevance to chemistry as they are the: tetrahedron, cube, octagon, icosahedron and dodecahedron and appear in crystal structures (tetrahedral and octahedral holes), molecular geometry and molecules (like cubane, SF$_6$). Topological work\cite{king1991topological} on polyhedral isomerisations describes the dynamic properties of the polyhedra for example a tetrahedron squashing to become a flat diamond, or a flat square distorting to become a diamond could have relevance in describing chemical reactions or organometallic compounds.

\subsubsection{Knot topology} Knot topology is a sub-field of algebraic topology and is the study of knots (see figure~\ref{fig:topol_example}) made by a circle of string, for example, the trefoil knot and Celtic knots (N.B. these are mathematical knots and they differ from those tied in string with two ends as used on boats or shoelaces), and has been the inspiration for several interesting chemical structures.\cite{fielden2017molecular} A sub-field of knot topology is link topology which is concerned with a collection of knots that do not intersect but can be linked together, also an inspiration for chemical structures for example, DNA links.\cite{baas2012chemica} 
%rotaxanes etc[[]]. 
Knots are classified and compared by being projected down to a 2-D space (imagine a shadow that a 3-D molecular structure model would cast on a wall) and various types of crossings and kinks are counted.\cite{sumners1987knot} 
%sumners1987knot - read this

Despite long molecules like proteins or DNA corresponding knot-topologically to the trivial and boring knot structure: the unknot (a 1-D string with two ends), chemists have expanded the mathematical idea of knot topology to approach the problems of looking at DNA and protein structure. For example\cite{adams2020knot} expands knot theory to knotoids or proteins having free ends, where overlaps and crossing give information about the structure as in knot topology.  I would argue that circuit topology is a version of knot topology. Circuits are made when intramolecular interactions like sulfur bridges are included giving a knot-topology less trivial than the unknot. Scalvini et al recently showed that these circuits are related to protein folding.\cite{scalvini2021topological} Although these methods are called topological by chemists, it has been suggested that this is really long range geometry.\cite{francl2009stretching} 
%[[All protein topology references here. ]]
Supercoiling of DNA to pack into small spaces is often referred to a topological property\cite{calladine1997understanding}, although it isn't a true mathematical topological property as uncoiling counts as a non destructive deformation. 

\subsubsection{Graph topology.} Most use of the word `topology' in chemistry refers to \textit{graph topology}, a different sub-field concerned with the connections of molecular graphs (i.e. chemical formulae), see figure~\ref{fig:topol_example}c, and when chemists use the word topological invariants they usually mean the invariants of those graphs or sub-graphs
\footnote{An invariant is something which is not changed under an operation.}. Topological invariants are those that are preserved under stretch, bend, deformation, are not measured, but should be quantitative, so we are left with counting, and these features have to be verifiable from within the object; we ca not `pull back' and look at the structure from outside.
The novel topological features presented in this paper do not come from graph topology on the molecular graph.

Network topology is to do with the shape of networks and invariants in their connections, and this has been used to try to define the chemical bonds from scalar fields and topological connections\footnote{Technically called bonds in the field of topology but as these do not always match up with chemical bonds the terminology leads to misunderstandings}.\cite{brown2002topology} Many molecular featurisations are based on the graph theoretic methods applied to molecular graphs and this necessarily includes graph topology. It has been suggested that molecular graphs are necessarily a 2-D description\cite{rouvray1995rationale}. however, by writing chemical formulae as 2-D graphs, one can use adjacency or connection matrices to create a topological graph invariant or a single number that defines the connectivity and is invariant to atom permutation, graph labelling and distance. Graphs can be analysed using spectral methods
%[[cite missing!]]
, and this has been used to predict biological properties,\cite{fabic1991study} boiling points\cite{basak1991predicting} inorganic structures\cite{brown2002topology} (this paper also includes an attempt to derive chemist-recognisable bonds from topology).

%quantum topology?

\subsubsection{Algebraic topology and her daughters persistence homology and topological data analysis}

\begin{figure*}[htp]
\small
  \begin{tabular*}{\textwidth}{@{\extracolsep{\fill}}ccc}
    \hline
    Structure & Point cloud & Persistence diagram\\
    \hline
    & & \\
 \includegraphics[width=4.5cm]{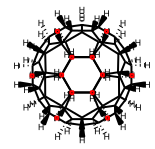} &
  \includegraphics[width=5cm]{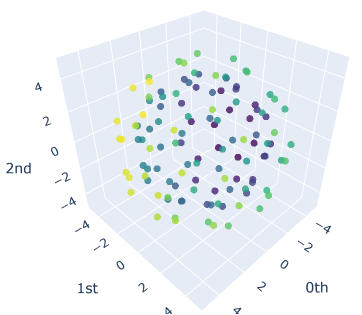} &
   \includegraphics[width=5cm]{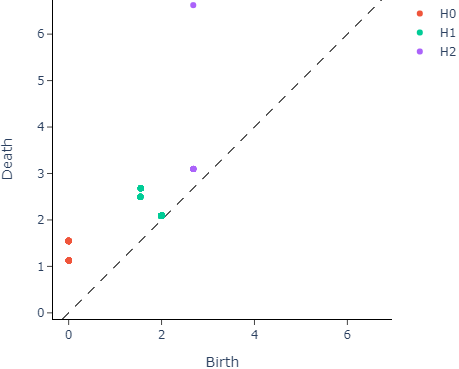}\\
 &
 
  \includegraphics[width=5cm]{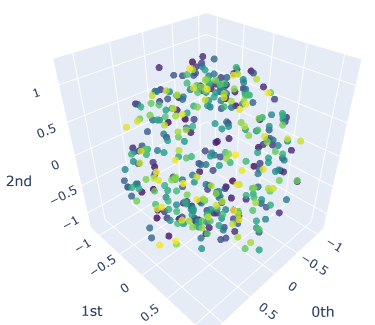}&
   \includegraphics[width=5cm]{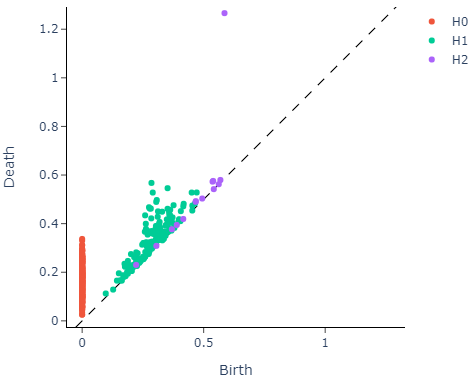}\\   
    \hline
  \end{tabular*}
   \caption{A buckyball is homologically similar to a sphere and a simplification of a noisy uniform spherical point distribution, or rather the crystalline version of a noisy sphere. As vibrations are added to such a model, it is clear that the persistence diagram will `distort' breaking the symmetry and multiplicity of the points and moving the diagram more towards the noisy sphere Top: left: Buckminsterfullerene structure; mid: Buckminsterfullerene point cloud; right: persistence diagram. Bottom: left and mid: a point cloud consisting of uniformly noisy points on a sphere, right: noisy sphere persistence diagram.} 
   %[[NTS: make a noisy spehre with 120 points for direct comparison]]}
 \label{fig:sphere_torus}
\end{figure*}

This is the area of topology mined for the novel molecular featurisation described in this paper. Although modern roots of the topology can be found back in 1700 with Euler, 
%(cite bridges problem and polyhedra) 
and the field really took off in 1900s, by mathematical terms, topology is one of the more recent mathematics to be discovered and is an area of active development by mathematicians, and as such, new results and sub-fields are emerging currently.  \emph{Topological data analysis} is a very new area of applied mathematics which is about applying algebraic and combinatorial topology techniques to data analysis, especially persistent homology, which is the extension of homology (\textit{vide infra}) to point clouds (such a atoms) and in which we are interested in features that persist over more than one length scale. Topological data analysis is the use of persistence homology to look at patterns in high dimensional data, see \cite{carlsson2009topology, wasserman2018topological, murugan2019introduction} and has been applied~\cite{pirashvili2018improved} to chemistry to look for patterns in the Delaney solubility dataset~\cite{delaney2004esol}. 

\subsection{What is algebraic topology?}
Formal mathematical descriptions are given in the appendix, here the concepts will be sketched in a non-formal and pictorial manner. Algebraic topology is concerned with comparing the shape of objects (in arbitrary numbers of dimensions), and homology is a method of comparing the shapes of `surfaces' of objects (technically, manifolds). Geometric relations, like the difference between rectilinear and curvilinear lines are not conserved, but \textit{connections} between parts are conserved, the number of holes or voids is used to classify types of shape. Measuring `holes', literally the lack of something, is difficult, and we have the restriction that we can not pull back from the object and look at it, we must use the properties of the surface to find its overall structure. This can be done by finding paths over the surface or building up \textit{covers} of the surface. Note that in topology a 3-D object's shape is a surface of the exteriour of that object. Classifications are made as to the number of holes in the object, such that a sphere and torus are different classifications of objects, and these classifications are given by the Betti numbers. For example, the surface of the wire-frame torus in figure~\ref{fig:topol_example}a has a single hole (although note that you could count the number of holes in the 3-D printed structure and get a much larger value). Objects are \textit{homotopically equivalent} if they have the same Betti numbers, For example, a tea-cup is homotopically equivalent to a donut  as both have a single hole, and if they were made of squashy, stretchy material, one can be smoothly manipulated to the other without breaking or tearing the material (see also figure~\ref{fig:math_concepts}).

%\cite{yoon2009mobius} mobius orbitals, real use of topoilogy - homology?

\subsection{Homology}

A manifold is a space similar to Euclidian space. Homology is the study of mappings of manifolds. We expect the spaces to be connected and allow them to be deformed as if they were made of stretchy, bendy material (topological deformation). Any property that is conserved under this sort of operation is a topological invariant. Cutting, sticking and gluing types of operations are not topologically invariant.\footnote{And a large part of topology is the construction of shapes by cutting, sticking and gluing.} The invariants that are preserved are things like the number of holes of any dimension. A hole of dimension 0 is a connection, so the number of connected regions are also topological invariants. 
The homology group $H_k(X)$ is a measure of the number of holes and connected regions of a space. The Betti numbers are related, as these are the rank of the homology groups and the number of cuts you can make in the object before it would be cut into separated regions. For example, a ring opening reaction cuts a molecule and changes its homology group, but does not stop the molecule from being connected. A hole is something that is not there, so it is hard to think about how it can be measured, and the solution is based on paths on the surface (you can remain on a surface and walk around a hole, but not across it, see figure~\ref{fig:math_concepts} in appendix~\ref{sec:maths}). 

\subsection{Persistence homology.} Persistence homology is the extension of homological ideas from surfaces to point clouds. As a point by definition has no volume, it is not obvious how a set of points might have a shape, volume and surface, despite the intuitive appeal of the idea. The solution is to use a balls of radius $\eta$ (and thus a volume in 3-D space) located on each point and examine that shape. It is rarely known which radius is the best to express the shape of the underlying datapoints, and the solution is to use balls of different radius. This approach has the advantage of mapping in some way the size of the homological features (connected regions and holes) as, for example, holes can become filled in as the ball radius grows. 

Homological features, $H_k$ are related to the dimension(s) of space that they are defined in. Connected regions are $H_0$ (1-D points). Flat 2-D holes are $H_1$, $k$ is 1, as holes in a 2-D surface are a 1-D feature in topology because topology is concerned with the surface and the surface of a hole is the line that defines its edge. Voids, for example, the void at the centre of a football, which can be thought of as holes in 3-D space are $H_2$ as the 3-D hole is defined by the 2-D surface that encapsulates it (the skin of the football). Homological features can continue up to arbitary numbers of dimensions, but in this paper we are only concerned with those features that can be found in 3-D space ($\mathbb{R}^3$).

\subsection{Homological features}

%[[change to table]]

\begin{table*}
\small
  \caption{Homological features ($H_k$) and their informal definitions. Chemical example is a molecule that has that homological feature as its main property.} %!!!add pictures}
  \label{tab:H}
  \begin{tabular*}{\textwidth}{@{\extracolsep{\fill}}lll}
    \hline
    Homological & Description  & Chemical example\\
    feature & &\\
    \hline
    $H_0$ & Connected region, no hole & Any long chain hydrocarbon \\
        &  & in textbook form \\
    $H_1$ & 1-D hole: found in 2-D space, like a hole punched in a piece of paper  & Benzene, heam (without iron)\\
   $H_2$ & 2-D hole: found in 3-D space, like the interiour of a football. & Buckminsterfullerene\\
 & N.B. a football is 2-D surface wrapped around a 2-D hold in topological terms  & \\
    $H_k, \: k>2$ & Higher dimensional voids. &N/A\\

    \hline
  \end{tabular*}
\end{table*}

%[[To add in, some figure of simpler chemical systems]]

\textit{Persistence} measures over which length scale a feature exists, for example, the large void at the centre of a bucky ball  persists over a large range of radii. A connected region is \textit{born} at radius 0 and \textit{die}s at the radius where that connected region joins with another one (and this approach will natural map out holes and voids as every object becomes a single connected component if the balls are large enough). A barcode plot plots the range of radii over which a connected region persists. A persistence diagram plots births against deaths for different homological features (see figure~\ref{fig:sphere_torus}), giving a rich set of information (metaphorically, one might consider this process a spectroscopy of shape). Bonds are included as the shortest bond to an atom will be the first connected feature to be born above a radius of 0, and as these merge distances between features corresponding to bonds and functional groups,see the red dots in figure~\ref{fig:sphere_torus}. However, the technique does not differentiate between formal chemical bonds and atoms close in space. I consider an advantage as bonds are a symbolic short-hand for the real situation and they sometimes obscure as much as enlighten. The $H_1$ feature can be seen to map the size of formal chemical rings, see the green dots in figure~\ref{fig:sphere_torus}. 
%as well as larger structures like the void (see the purple dot far off the line in figure~\ref{fig:sphere_torus}). 
The final feature to close gives the topological class of the object, so in figure~\ref{fig:sphere_torus} we see that the bucky ball is homologically equivalent to a sphere with the purple $H_2$ feature far off the line. Interestingly, mathematicians studying persistence homology of multi-dimensional datasets believe (but have not proved) that features that do not persist over long scales to be noise or irrelevant to the structure of the dataset.\cite{atienza2016separating} In these examples, we find that the features that do not persist for long are, at small radii, related to the incredibly chemically important features of bonds, angles, torsions and functional groups.
%¬!!! NTS - there were more references discussing whether unpersistent features are noise or not

Features based on persistent homology encode a rough shape, much like a space-filling model\footnote{Aside: in the language of topology, space-filling models are dual to ball and stick models, as balls are 0-simplices), sticks are  1-simplices and space filling models are made of 3-simplices (volumes).} however they do not include atom types in the shape and so `ignore' chirality\footnote{The lack of atomic labels makes it hard to differentiate between chiral entantomers, but bond length differences are preserved, so chirality is perhaps not completely absent.} but preserve the difference between isomers. I am also developing a 3-D input method based on the tools of geometric machine learning which preserves chirality (paper forthcoming) and combining these input types is likely to be fruitful in further featurisation development. 
%[[cite patent]]

A key idea in modern mathematics dating to a lecture by Riemann given in 1854 is the every mathematical object parameterized by $n$ real numbers can be treated as a point in $n$-dimensional space.
%[[4]]:140. 
Graph topology can be considered the \textit{`Euclidization of data'}\cite{saveliev2016topology} which moves the data points into a 3-D space to investigate the topology. Molecular graphs are a 2-D representation of a molecule's formal bond structure. As chemists, we often use wedges or position to encode some aspect of a molecule's structure and these chemical formulae might be considered (informally and non-mathemtically) 2 and a bit dimensions, however it should be noted that it is our human perception and chemical knowledge that fills in this information: an ML algorithm does not get that from the graph.\footnote{This may not be entirely true, it is possible that given enough data an ML algorithm might be learning something about 3-D space from these 2-D representations.}  The drug in figure~\ref{fig:topol_example} is drawn missing its six stereocentres to show what the actual molecular graph is. A topological analysis of a molecular graph then moves this 2-D formula into Euclidean space and preserves only the 2-D structure and connections between atoms and not the distances or angles (as these are the province of geometry). The technique presented herein instead \textit{`discretizes Euclidean space'}, specifically 3-D space, then creates a graph based on the shape and calculates topological features, i.e. the 3-D structure is used as an input, and due to using persistent homology, some small treatment of distance, and thus geometry, is preserved. The relation between Euclidean and topological spaces is forgetful, so information is lost during the transform and it is `impossible'\footnote{This is the inverse problem. I suspect that it is not impossible, but would require guessing a molecular structure based on topological features and refinement of those guesses.} to convert from the topological features back to the molecule directly. 

In this paper, we will use the tools of persistence homology to study and encode 3-D chemical structure. With the best attempt at a literature survey, I believe this is the first time these tools have been applied to these specific problems. There are many ways of inputting chemical structure into a computer, including some based on graph topology. I posit that 3-D structure is essential for some chemical learning tasks.
%[[cite plus drawbacks]]. 
The persistence homology encodings used here will encode shape properties for use in ML. The resulting persistence diagrams give us much information about the topological properties of a shape. We will `distill down' this information by vectorizing and embedding the persistence diagrams into vectors of length no more than 18 per molecule, making the smallest 3-D chemical structure encoding yet found. With these novel chemical structure encodings, we can ask the question as to which problems are mostly dependent on 3-D structure, whether the information encoded in this approach is sufficient, or when might we need to include extra information. A study of what extra information is needed for each task is included, with a focus on cheap and easy to calculate non-topological features.

\subsubsection{Molecular machine learning problems attempted\label{ssec:chem_prob}}

We try our method on several standard molecule ML problems, see table~\ref{tbl:datasets}, with a focus towards data that is necessary for drug development, like solubility and toxicity. 

\begin{table*}
\small
  \caption{\ Datasets investigated.}
  \label{tbl:datasets}
  \begin{tabular*}{\textwidth}{@{\extracolsep{\fill}}llll}
    \hline
    Dataset & Problem area & Problem type & Target \\
    \hline
%    BBBP & Physiology & Classification & Penetration of the Blood-brain barrier penetration \\
    %PDBBIND & Protein-ligand binding & Regression & Binding affinities of biomolecular complexes \\
    QM7 & Quantum mechanics & Regression & Atomisation energy\\
    QM8 & Quantum mechanics & Regression & Electronic spectra and excited \\
    &  &  &  state energy of small molecules.\\
    Delaney & Physical chemistry & Regression & Water solubility data\\
    &  &  &  for common organic small molecules\\
    FreeSolv & Physical Chemistry & Regression & Experimental and calculated hydration free energy of small molecules in water.\\
    Lipophilicity & Drug development & Regression & Experimental results of octanol/water distribution coefficient(logD at pH 7.4).\\
    Tox21 & Drug development & Classification & Qualitative toxicity measurements on 12 biological targets\\
    \hline
  \end{tabular*}
\end{table*}

\section{Methodology\label{sec:methodology}}

\subsection{Creating persistence homology topological indicators with example simple molecules}

To make the concepts discussed in this paper clearer, we shall briefly survey the persistence diagrams of some small molecules. 

\subsection{Point clouds}

First, we turn atom positions to point clouds, this removes atom type, formal bonds, atom size, valence, mass and charge, leaving only shape information. The only chemical information left is that which could be inferred from the distances between atoms by the ML algorithm.

\subsection{Simplices}

\begin{figure}
    \centering
    \includegraphics[width=0.45\textwidth]{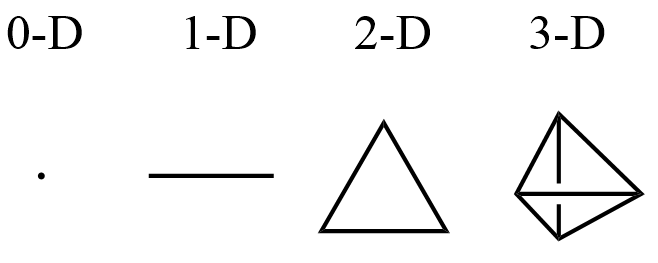}
    \caption{0-, 1-, 2- and 3-D simplices are very familiar to chemists.}
    \label{fig:simplices}
\end{figure}

It might seem strange, given that our interest is in curved surfaces, which are the underlying smoothed `shape' of a molecule, the next step is to use straight and flat objects to quantify this shape. This is because topology does not differentiate between curved and straight lines, and we are only interested in the connections between parts, thus we make the task simple and use simple objects called simplices to quantify this connection.

3-D simplices are very familiar to chemists as they are tetrahedrons, the concept can be extended to any dimensionality of space, however only four concern us here: the point (0-D), the line (1-D), the triangle (2-D) and the tetrahedron (3-D), see figure~\ref{fig:simplices}. Note that simplices can have parts of any length, they do not need to be equal as drawn in figure~\ref{fig:simplices}. A polytope is a geometric object with `flat' sides and a simplex is the simplest possible polytope in any given space. For example, a triangle is a flat 2-D object with three sides, and adding more sides makes a less simple polytope, removing one side makes two 1-D lines (i.e. not a polytope). The best way to think of `flatness' is that the tangents to the $n$-simplex is are parallel in the $n+1$ dimension. i.e. a line in $x$ has parallel tangents all along it in $y$, a triangle in 2-D space $(x,y)$ has parallel tangents all along the face in $z$ and so on (although note that we must work in 2-D space for this construction to work), see also appendix~\ref{ssec:flatness}.

\begin{table}[htp]
    \centering
    \begin{tabular}{ccccc}
    \hline
        Manifold    & Dimemsion & $\beta_0$ & $\beta_1$ & $\beta_2$ \\
    \hline
         Circle, $S^1$& 1-D & 1 & 1 & -\\
         Sphere, $S^2$& 2-D & 1 & 0 &  1 \\
         Torus, $T^2$& 2-D  & 1 & 2 & 1\\
         \hline

    \end{tabular}
    \caption{Example topological surfaces (manifolds) and their Betti numbers. N.B. The circle is the 2-D version of the sphere, thus the boundary of the circle is the 1-D version of the surface of the sphere.}
    \label{tab:shapes}
\end{table}

\begin{table*}[htp]
    \centering
    \begin{tabular}{lp{12cm}}
    Term & Informal definition \\
    \hline

    Topological invariant & Properties that remain constant under topological deformation.\\
    Homeotopically equivalent  & Same topological shape, one can be continuously deformed into the other, i.e. you can map $X$ to $Y$.\\
    Homeomorphically equivalent & Like homotopically equivalent, but you can deform $X$ to $Y$ and $Y$ to $X$.\\
    Connected space & Not a union of two disjoint non-empty open sets. Benzene is not a single connected space in simplical complexs $X_0$ and $X_1$ in figure~\ref{fig:benzene_simplex}, it is a connected space in simplical complexes $X_2$, $X_3$, $X_4$ and $X_5$.\\
    $H_k (X)$ & Homology group $k$ of $X$, generates paths around holes of dimension $k$.\\
    %Homology group/generator $H_k$? & \\ %nested simplical complexes \\
    Betti numbers $\beta_{k}$ &  Rank of $H_k$, the number of cuts that can be made before $X$ is no longer connected.\\
    $\beta_0$ & Number of connected components. \\
    $\beta_1$ & Number of 1-D `flat' holes, \\
    $\beta_2$ & Number of 2-D holes (voids). \\
    \end{tabular}
    \caption{Terminology}
    \label{tab:terminology}
\end{table*}

%Betti numbers tell you the maximum number of cuts that can be made in a surface before separating it into two peices
%a 0-dimensional boundary is the gap between two components, so $H_0 (X)$ is the path connected components of X 

%short persistence is what mathematicians call noise

\subsection{Building simplical complexes}

If a ball of radius $\eta$ is centred on two atoms, now modeled as points, and the radius of that ball is half the distance between them such that the balls touch or overlap, those points are now connected. In chemistry, this generally corresponds to a formal bond. The next step of the calculation would be to draw a 1-simplex (straight line) between the two 0-simplices (points) that describe the object. For three atoms, we draw a triangle between them if the balls overlap. simplices are hierarchical, so a triangle consists of three 0-simplices (points), three 1-simplices (lines) and one 2-simplex (triangle). Ditto a tetrahedron four 0-simplices (points), six 1-simplices (lines) and four 2-simplices (triangle) and one 3-simplex (tetrahedron)\footnote{A. Higher order simplices can exist in higher dimensional space. B. You can create cubical complexes from points, lines, squares and cubes, but these are less simple as they have more lower-order components.}. Note that the torus presented in figure~\ref{fig:topol_example} is actually a simplical complex built on a torus using line simplices.

For a threshold $\eta$ the procedure is as follows:
\begin{enumerate}
    \item add a line (1-simplex) between any two points that are less than $\eta$ from each other (i.e. balls overlap)
    \item add a triangle (2-simplex) between any three points that are less than $\eta$ from each other
    \item add a tetrahedron (3-simplex) between any four points that are less than $\eta$ from each other.
\end{enumerate}

The result is a simplical complex (a connected object made up of simplices).

We write simplices as the points that define them, so a line between $a$ and $b$ is written as $[a, b]$, and this is actually the set of all points between $a$ and $b$ that satisfy the constraints for `flatness' (see also appendix~\ref{ssec:simplical_complex}). 

\subsection{Creation of Vectoris-Rips complex}

A series of simplical complexes are created at different ball sizes and these are combined into a heirachical complex called a Vectoris-Rips complex. 

\subsection{Barcodes and perisistence diagrams}

The filtration at which a generator appears is called birth, that at which is dies is called death, and 
$$
persistence = death - birth
$$
And we can plot the persistence of generators in persistence diagrams (See figures~\ref{fig:sphere_torus} and ~\ref{fig:6_C}), barcode plots or Betti curves (see figure~\ref{fig:benzene_barcode}).

Persistence diagrams are a compact and easy way to represent the information in a barcode plot (especially when you have many points) as the filtration is changed, i.e. the balls diameter is increased. Births refers to the creation of a feature, deaths when that features dies (by merging with another feature). Points on the birth-death line would be points created and then immediately destroyed, and many of the points on this line correspond to non-persistent features and are designated as noise.\cite{atienza2016separating} The more persistent a feature, the more important it is considered. Those points that persist over a range of filtrations are high off the line, and we read the diagrams by looking at these to find out what homology group generators are present (i.e. how many clusters, holes and voids).

\subsection{Vectorisation of persistence diagrams}

Persistence diagrams are easy for humans to interpret but not very useful for inputting into a machine learning algorithm. To do that we use five features derived from the persistence diagram:
\begin{enumerate}
    \item Persistence entropy\cite{rucco2016characterisation},
    \item No. of points, 
    \item Bottleneck distance,\cite{kerber2017geometry}
    \item Wasserstein amplitude,\cite{kerber2017geometry}
    \item Persistence image,\cite{adams2017persistence} 
\end{enumerate}
and each feature is 3-D giving an input vector of 18 numbers long for each input. A description of how these features are calculated is in the appendix~\ref{sec:maths}. 

\setcounter{footnote}{0}

This input format is very information dense and excellent for machine learning. There are two problems with inputs for MML, often the input vectors are sparse (i.e. they contain lots of zeros), and the input vector has to be the same for each molecule regardless of the molecule's size and complexity, and this is usually dealt with by taking the largest molecule in the dataset to set the input size and this also increases the sparsity. Neural networks, for example, find it very hard to learn from sparse datasets.\cite{gale2018and}

\begin{figure}[htbp]
\centering
  \includegraphics[height=4cm]{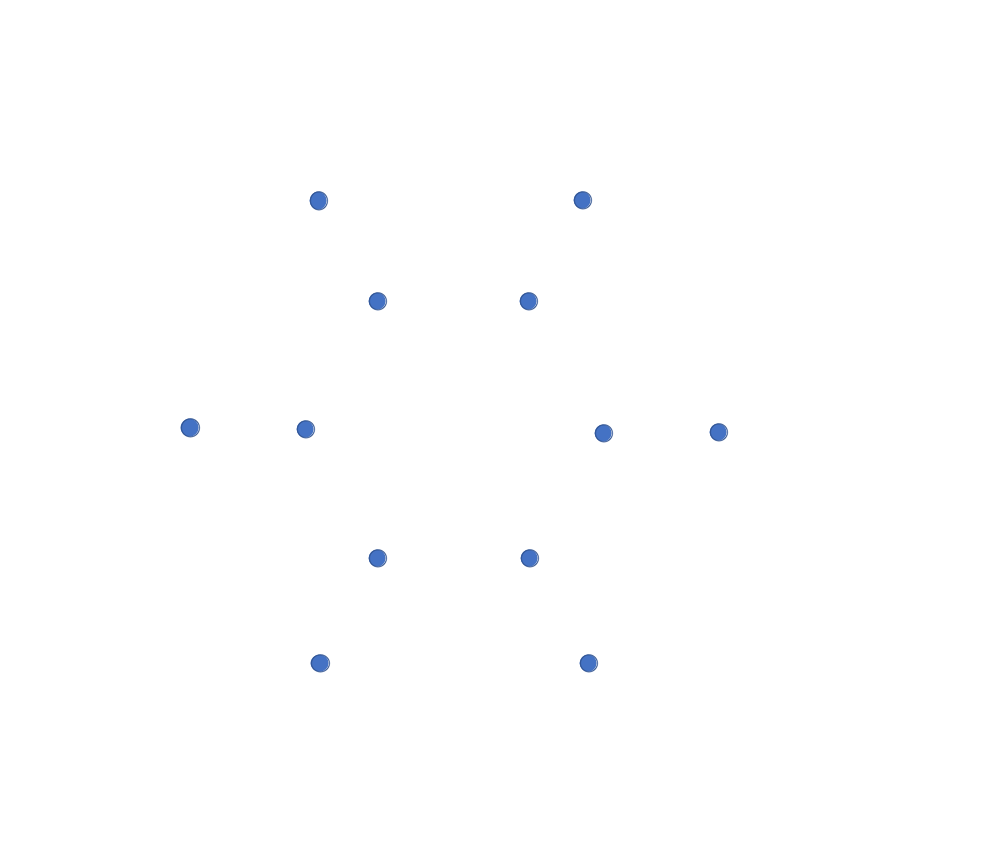}\\
  a. $X_0$ is 12 simplex-0 (points), $\beta_0=12$, $\beta_1=0$, $\beta_2=0$, class: cluster
  \\
  \includegraphics[height=4cm]{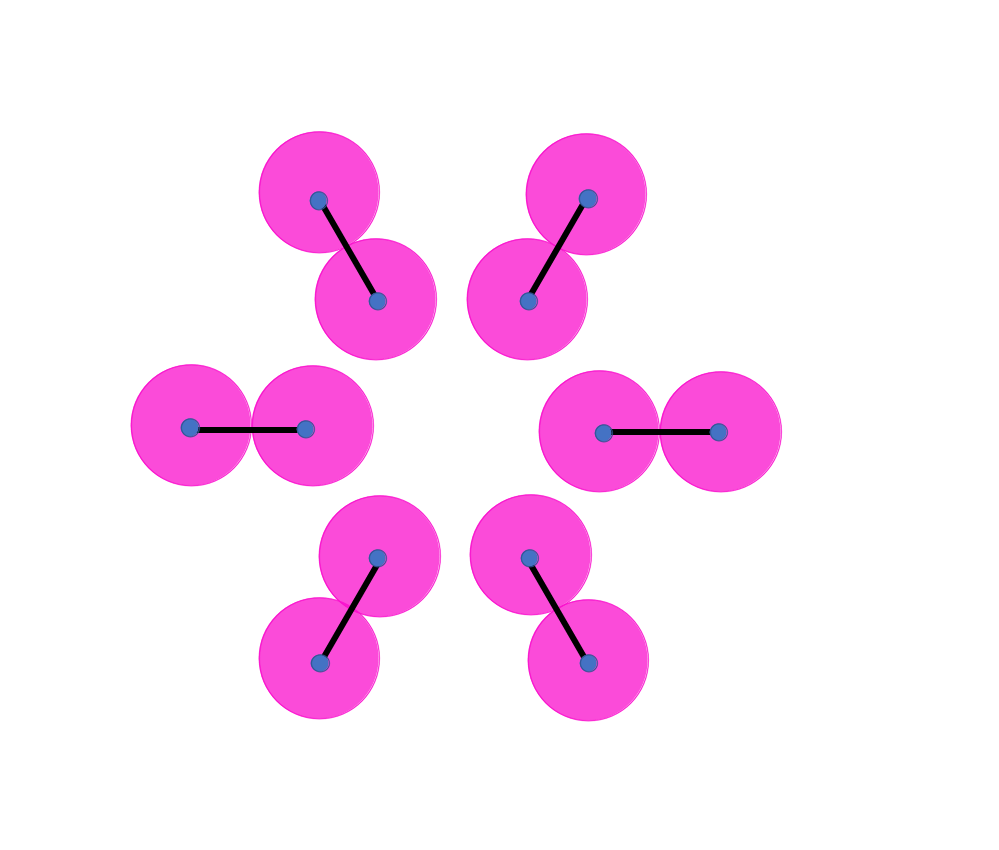}\\
  b. $X_1$: 6 simplex-1 added , $\beta_0=6$, $\beta_1=0$, $\beta_2=0$, class: cluster\\
  \includegraphics[height=4cm]{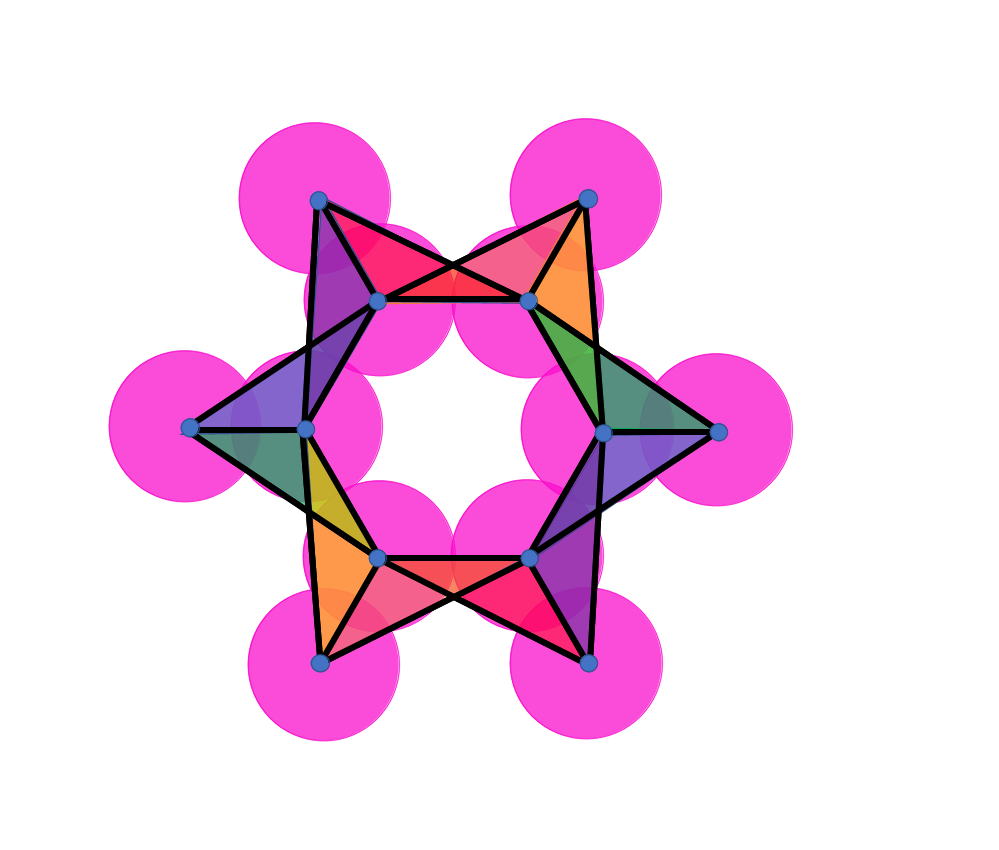}\\
  $X_2$: 6 simplex-1 added, 12 simplex-2 added , $\beta_0=1$, $\beta_1=1$, $\beta_2=0$, class: circle\\
  \includegraphics[height=4cm]{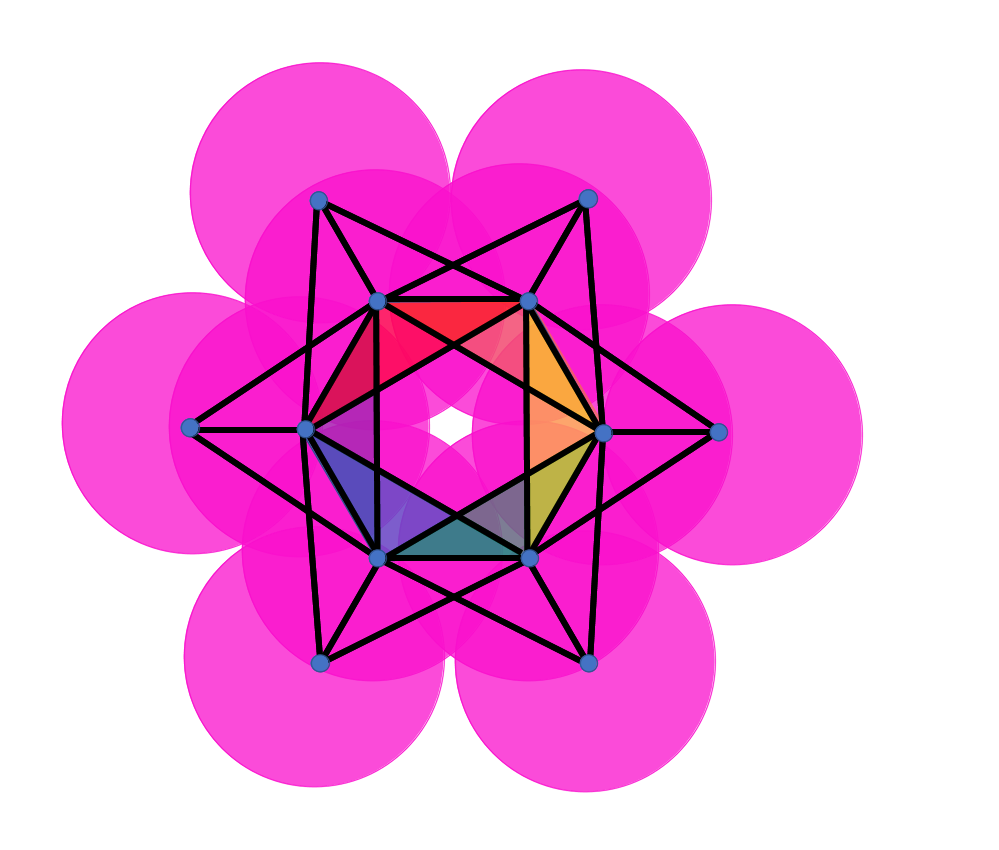}\\
  c. $X_3$: 6 simplex-2 added, $\beta_0=1$, $\beta_1=1$, $\beta_2=0$, class: circle\\
  \includegraphics[height=4cm]{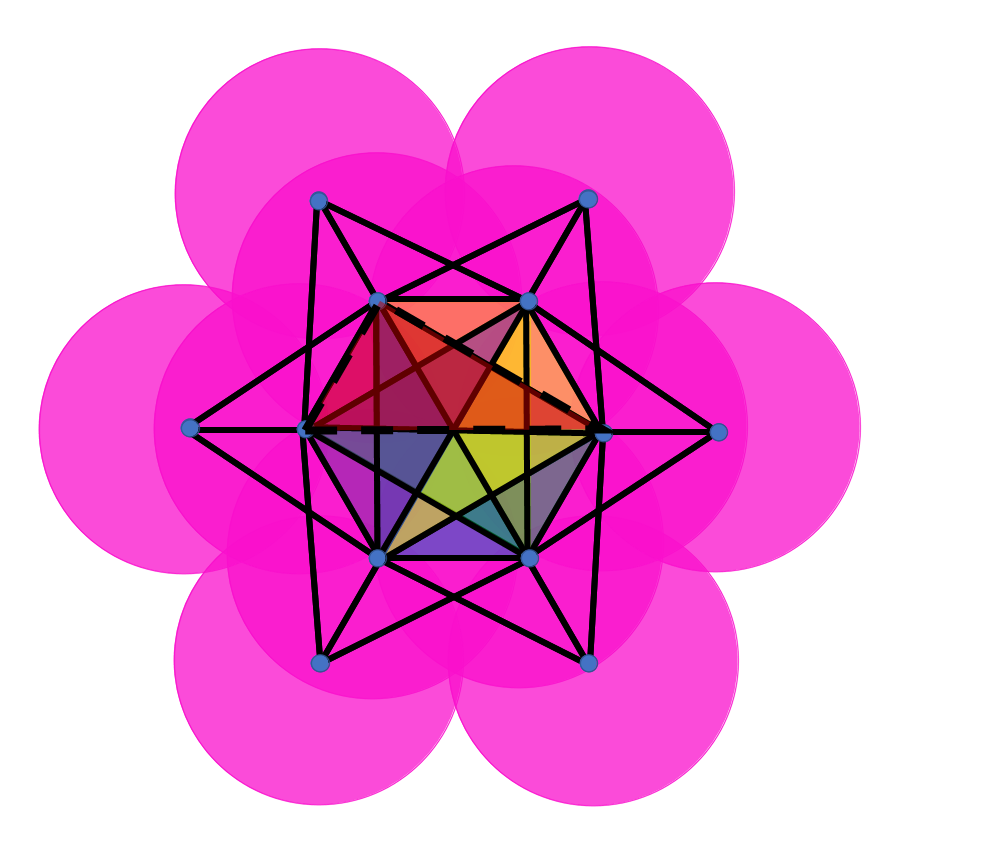}\\
  e. $X_4$ , $\beta_0=1$, $\beta_1=0$, $\beta_2=0$, class cluster\\
  \caption{How simplical complexes are built up with increasing ball size.}
  \label{fig:benzene_simplex}
\end{figure}

\subsection{Computational details}

Datasets were taken from MoleculeNet\cite{wu2018moleculenet}, were featurised with the available featurisations, ECFP, grid, graph convolution, MACCS, rdkit, Coulomb matrix and the eigenvalues of the Coulomb matrix (CM\_eig) in DeepChem. Molecular structures were created and minimised using rdkit.\cite{rdkit} PHF features were created using the Giotto-TDA topological data analysis module\cite{Giotto-tda,tauzin2020giotto}. Output TDA features were saved as hdf5 files. All features were trained in DeepChem using the standard benchmark models\cite{wu2018moleculenet} suitable for each feature, i.e. PHF and other 1-D features were fed into 2 layer, feed-forward neural networks with 1000 units in each layer. Graph featurisations were fed into graph convolutional neural networks. All control experiments were performed using the set-up and models in DeepChem v2.3 as described in the MoleculeNet paper\cite{wu2018moleculenet}.

\subsection{Software}

All code is given in GitHub project `Graphs and topology for chemistry'\cite{GandT} and was created building on Giotta-tda\cite{Giotto-tda,tauzin2020giotto} and rdkit\cite{rdkit}. Example code is given in appendix~\ref{sec:code}. The package contains code to featurise molecular structures, along with scripts to featurise the datasets given in table~\ref{tab:datasets}. To run the code as in the MoleculeNet paper\cite{wu2018moleculenet}, you will need DeepChem 2.3\cite{wu2018moleculenet}, a forked version of this along with verification scripts is given in\cite{DCval} An updated version using DeepChem 2.6 is also provided.\footnote{There were large changes in the underlying code between DeepChem 2.3 and 2.6.}
%% !!! the datasets are available from ...

\section{Benzene example \label{sec:benzene}}

To make the concepts in section~\ref{sec:methodology} clear and to investigate what information is preserved by persistent homology featurisation we will now go through the method applied to a very simple molecule: benzene.

\subsubsection{Creating Vectoris-Rips complex.}

First we make a point cloud of the atoms as shown in figure~\ref{fig:benzene_simplex}a. We then increase the value of $\eta$ to make a series of complexes and record where homological groups ($H_k$) are born and die. Betti numbers $\beta_k$ are the number of features of a given homological type. 

We record all the simplical features in complexes $X_n$.

If $\eta$ is less that $^{r_{CH}}/_2$, we only have points:
$$
X_0 = \{C_1, C2, ... C_6, H_1, H_2, ... H_6 \}
$$
and the Betti numbers are $\beta_0 = 12$, $\beta_1 = 0$, $\beta_2 = 0$, i.e. there are 12 connected components and the complex $X_0$ consists of a set of points. Here we have labelled them with the atoms for ease of understanding but in the method these are equivalent.

If $\eta$ is more than $^{r_{CH}}/_2$ but less than $^{r_{CC}}/_2$, 1-simplices are formed between the C-H atoms: we have 12 points (simplex-0) and 6 lines (simplex-1):
$$
X_1 = \{X_0, [C_1, H_1], ... [C_6, H_6],
$$
as we now have 6 connected components (clusters) (that correspond to teh cluster of C and H points in a CH bond), see figure~\ref{fig:benzene_simplex}c
%[[check betti numbers]]
, thus $\beta_0 = 6$, $\beta_1 = 0$, $\beta_2 = 0$, topologically this is 6 clusters.

If $\eta$ is more than $r_{CC} / 2$ we have 12 points and 12 lines and 12 triangles:
$$
X_2 = \{X_0, X_1, [C_1, C_2], ... [C_6, C_1], 
$$
$$
[H_1, C_1, C_2], [H_1, C_1, C_6], ... 
$$
$$
[H_6, C_6, C_5], [H_6, C_6, C_1]\} \;,
$$
The CCH triangles overlap creating a loop, see figure~\ref{fig:benzene_simplex} as this is a 1-D surface we now know that we have a 2-D hole in this structure, and the Betti number, $\beta_1 = 1$. So $\beta_0 = 1$, $\beta_1 = 1$, $\beta_2 = 0$. At this point benzene is homotopically equivalent to a circle ($S^1$), see table~\ref{tab:shapes}. This is conceptually slightly odd for a chemist as the benzene ring structure has been built from HCH triangles overlapping rather than the ring of CC bonds. 

Increasing $\eta$, we add in the CCC angles as 2-simplices:
$$
X_3 = \{X_0, X_1, X_2, [C_1, C_2, C_3], ... [C_6, C_1, C_2]\}
$$
see figure~\ref{fig:benzene_simplex}d.
As the topology was already that of a circle, topologically nothing is changed by this addition. We have 12 points, 24 lines and 18 triangles.

\begin{figure}
    \centering
    \includegraphics{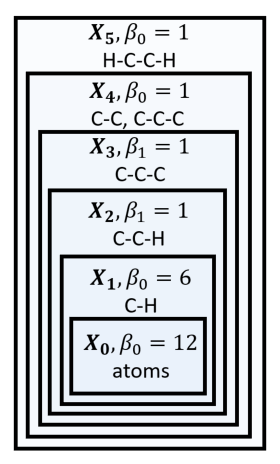}
    \caption{Vectoris-Rips complex for benzene. The complex is hierarchical and compositional; higher values of $\eta$ give complexes composed of the simplices at that ballsize and the simplical complexes found at smaller radii.Only the largest non-zero Betti numbers are shown.}
    \label{fig:VR_complex}
\end{figure}

When $\eta$ is half the diameter of a benzene molecule, we add connections across the ring, and new triangles:
$$
X_4 = \{X_0, X_1, X_2, X_3, [C_1, C_4], ... [C_3, C_6], [C_1, C_3, C_4], [C_6, C_2, C_3]\}
$$
and the Betti numbers are $\beta_1 = 1$, $\beta_1 = 0$, $\beta_2 = 1$ and the loop dies, 

Eventually, the hydrogens across the ring join:
$$
X_5 = \{X_0, X_1, X_2, X_3, X_4, [H_1, C_1, C_4, H_4], ... [H_6, C_6, C_3, H_3]\}
$$
and everything is connected, and benzene is now homotopically equivalent to a sphere and is a single cluster, see figure~\ref{fig:benzene_simplex}e, $\beta_0 = 1$, $\beta_1 = 0$, $\beta_2 = 0$ As this join can be thought of as going over the top, the hole(s) that are joined in 3-D. If you think about the hydrogens being slightly bent up and out of the plane it is obvious that a 3-D void would be formed above the plane of the carbons. The rdkit version is only flat to five decimals places (in fact the six C-H bonds did not have the exact same value but this cannot be seen in the persistence plot as they overlap), which can explain the presence of the void.\footnote{Although, if the benzene is perfectly flat then from the reasoning of the amount the hydrogens are above the plane tending to zero there is an infinitesimal point where the void appears and disappears. This would appear on the birth=death line.} We expect that this example has also made clear the links between topology and combinatorics. The final result is the Veirtoris-Rips complex, which is a set of sets of complexes, as illustrated in figure~\ref{fig:VR_complex}.

From the benzene example, it is tempting to relate the simplices to something a chemistry would would recognise and name, i.e bonds, angles and torsions, thinking this way is problematic for two reasons. Firstly, angles and torsions are geometric properties, expecting straight lines and measurable angles. Topology does not measure or preserve angles, straight-lines are equivalent to curves,\footnote{We're using straight lines and flat shapes for convenience, and they are the simplest way to draw connections, hence simplices.} and there are no atom labels being used, so when we write [C, H, H] it is just connection between three points in space. Secondly, not all the connected parts are something a chemist would recognise and label. For example, the hydrogens above cyclohexane in a boat configuration will be joined by simplices but are five bonds away (see figure~\ref{fig:6_C}). Nearby atoms on neighbouring proteins chains would also be joined. Counter ions also. If this topological approach works for a problem we can assume that these longer range properties are required to solve the problem.

\subsubsection{Creating the barcode plot}

Figure~\ref{fig:VR_complex} shows the final Veirtoris-Rips complex as a heirarchical set of sets of simplices relating to discrete values of $\eta$, however the topological features exist over a continuum of lengthscales. We are interested in the persistence of these homological generators at different lengthscales, which is what is plotted against a continuous scale in the barcode plot. 
\setcounter{footnote}{0}

\begin{figure}[htbp]
\centering
  \includegraphics[width=6cm]{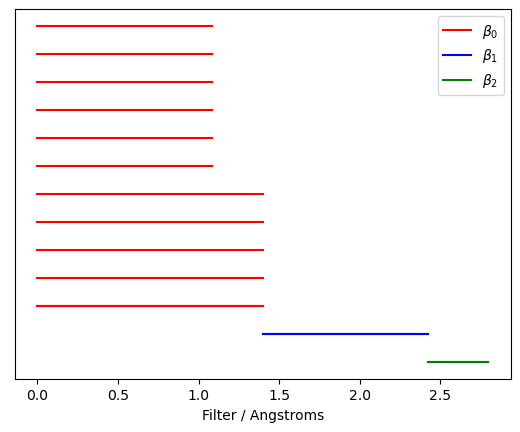}\\
  \includegraphics[width=6.5cm]{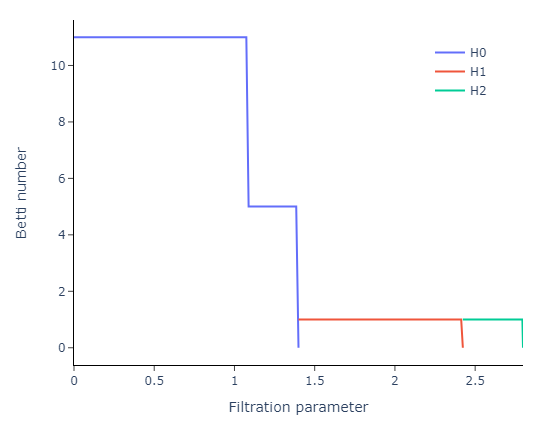}\\
  \caption{The barcode plot (top) and Betti curve (bottom) for benzene.}
  \label{fig:benzene_barcode}
\end{figure}

A barcode plot records the dramatically named birth and death of homological features, see figure~\ref{fig:benzene_barcode}, over the length scale (called filter in this field as we are filtering the simplical complexes).  There are 12 points (simplex-0) in benzene, there is always a point feature that continues to infinity (which we do not plot), so there are only 11 points on the barcode plot in \ref{fig:benzene_barcode}. The 12 connected regions are born at 0\AA $\,$ and coloured red on the diagram. Six of them die at the C-H bond length when they merge with the other six to create 6 clusters.  Note that the atoms are not numbered or differentiated, so there is no difference between them, and the ordering over the barcode plot is merely for convenience.\footnote{It might be interesting to plot the birth and death of these features ordered as distance from a point to give a spatial ordering over the features, but this not pure topology and not investigated in this paper.} Six of these simplex-0 die at 1.08\AA, which is the C-H bond distance, and this corresponds to figure ~\ref{fig:benzene_simplex}b and simplical complex $X_1$. At a filter of 1.4\AA, the C-C bond length, the last remaining clusters die and a new feature, of homology group $H_1$ is born, i.e. the clusters merge to become a 2-D hole with a 1-D surface around it (the carbon ring), this is simplical complex $X_2$, see figure~\ref{fig:benzene_simplex}c. This ring persists for 1.024\AA~ before it closes (simplical complex $X_4$ and figure~\ref{fig:benzene_simplex}e). At this filter value, benzene is indistinguishable from a single cluster, and any increases in ball size will not change the homology of the complex. This single cluster persists until $\infty$\AA, but this final point is not plotted on the barcode plot, and this explains why there are only 11 features in figure~\ref{fig:benzene_barcode}. 

\subsubsection{Persistence diagram}

Each topological feature is described by a pair of numbers, $\eta$, $t$ where $\eta$ is the current ball radius and $t$ is the maximum radius achieved. 
%[[sort this(i think)]]. 
This is converted to a multiplicity of points $(b, d)$ (see appendix~\ref{sec:maths} where $b$ is the radius where the feature is born and $d$ is the radius where is dies. The persistence diagram, despite being called a diagram, is actually the multiset of points as shown for benzene in table ~\ref{tab:persistence}. Each topological feature has three numbers associated with it, the radius at which it is born, the radius at which it dies and the Betti number for that feature. There is a final feature that is often excluded from persistence diagrams and that is the feature that starts at the point that the penultimate feature dies (the last row in table~\ref{tab:persistence}) and dies at infinity--this is not plotted.\footnote{This has a nice metaphorical interpretation which fits with our understanding of the human visual system, at large distance, every shape becomes a point.} For molecules, there will always be $N-1$ original features for this reason (notice that one of the hydrogens are missing from the barcode plot).  

The data in the persistence diagram in table~\ref{tab:persistence} can be plotted on something which is confusingly also called a persistence diagram, and examples of these are shown in figures~\ref{fig:sphere_torus} and ~\ref{fig:6_C}. These persistence diagrams plot the data as multisets of points $(b, d)$ i.e. we plot birth against death with the Betti number (third column) shown as colour.
%(multisets defined in appendix[[]]). 
If there are several points with the same $(b, d)$ they are plotted on top of each other. In the software, hovering a mouse over the points reveals the multiplicity of that value, one could also annotate the diagram with the number next to it. Obviously, molecules with a high degree of symmetry and those in the perfect ground structure will have persistence diagrams with a high degree of multiplicity. As molecules move, or the structure changes slightly these multiple points move apart, as shown in in figure~\ref{fig:6_C}. 

\begin{table}[htp]
\small
  \caption{Persistence diagram for benzene}
  \label{tab:persistence}
  \begin{tabular*}{0.48\textwidth}{@{\extracolsep{\fill}}lll}
    \hline
    Birth & Death & $k$ in $H_k$ \\
    \hline
    0.        & 1.08233893 & 0.        \\
    0.        &  1.08233905 & 0.      \\
    0.        &  1.08233905 &  0.   \\
    0.        &  1.08233905 & 0.    \\
    0.        &  1.08233905 &  0.    \\
    0.        & 1.08233905 & 0.     \\
    0.        &  1.39904296 & 0.      \\
    0.        &  1.39904296 & 0.    \\
    0.        &  1.39904296 &  0.     \\
    0.        &  1.39904296 & 0.    \\
    0.        &  1.39904296 & 0.     \\
    1.39904296 & 2.42321348 &  1.    \\
    2.42321348 &  2.79808593 & 2.     \\
    \hline
  \end{tabular*}
\end{table}

What is the homology class of a benzene molecule? This is where persistence is useful, we choose to discard features that are not persistent. There is not a formal mathematical rule for doing this. Generally we might approximate it as the highest ranking $H_0$ with the largest persistence, and from the data above, that would suggest that benzene is homotopically equivalent to a circle, i.e. it has one (flat) hole. The connected regions ($H_0$) also persist for a long time, which tells us that this larger hole structure is made of clusters (which makes chemical sense). In topological data analysis the closer the point is to the line, the more likely it is to be noise. In these chemical structures these `noisy' points seem to encode something interesting about the structure. For example, in benzene, the $H_2$ could be ignored as noise, but it does make sense in terms of benzene being a 3-D structure and not being entirely flat, which could lead to the fleeting existence of a void above or below the molecule.

The process for creating these topological features may seem rather involved, however, it is important to note that this can be done in only four lines of python code (see Appendix \ref{sec:code_example}). The interested reader is referred to Appendix~\ref{sec:maths} for further, more formal mathematical detail.

\section{Other simple molecules}

%%% expand on this later hwne you compare different molecules and use this as a simularity measure
%'A well known result establishes that there exists an isomorphism between two persistence module if and only if their persistence diagrams are equal.' from giotta

%think this means that molecule shapes are equal only if their persistence diagrams are equal. But you can talk about similarity - I think - put htis in when discussing figs of spheres and bucky balls and c6 stuff. 

\begin{figure*}
\small
  \begin{tabular*}{\textwidth}{@{\extracolsep{\fill}}ccc}
    \hline
    Structure & Point cloud & Persistence diagram\\
    \hline
 \includegraphics[width=3cm]{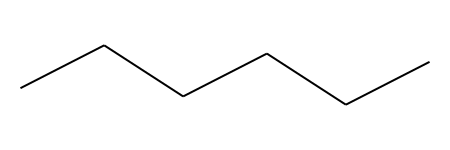} &
  \includegraphics[width=5cm]{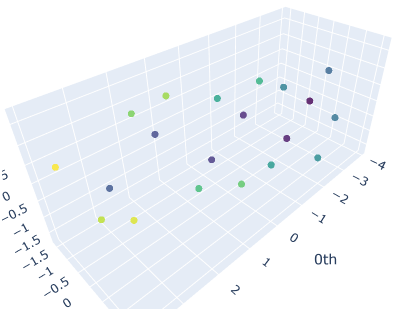} &
   \includegraphics[height=5.5cm]{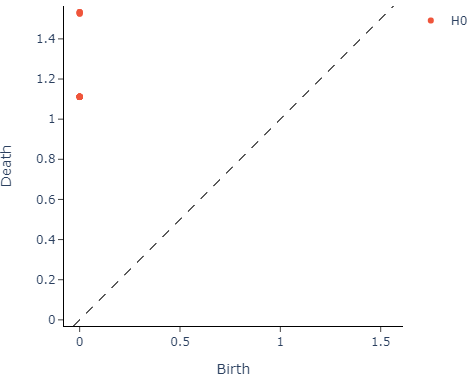}\\
 \includegraphics[width=2cm]{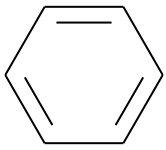} &
  \includegraphics[width=5cm]{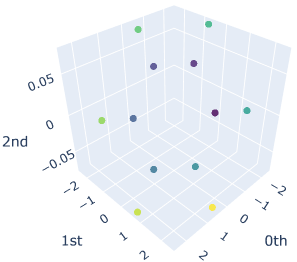} &
   \includegraphics[height=5.5cm]{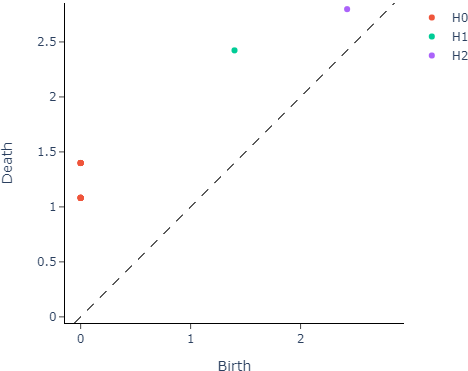}\\
 \includegraphics[width=3cm]{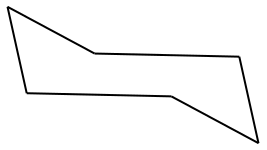} &
  \includegraphics[width=5cm]{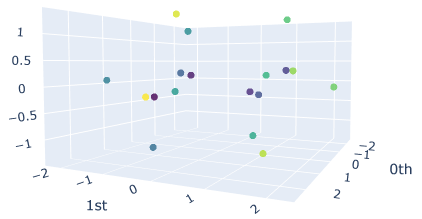} &
   \includegraphics[height=5.5cm]{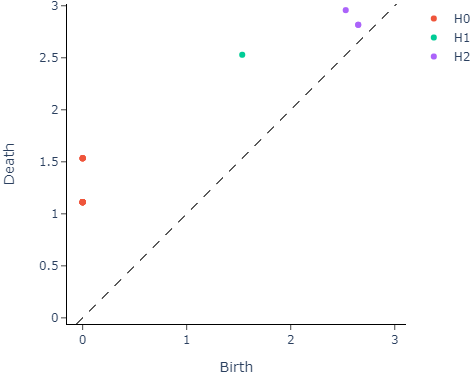}\\
  \includegraphics[width=3cm]{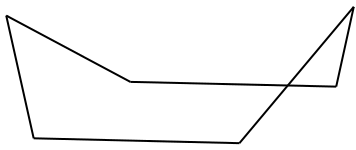}&
   \includegraphics[width=5cm]{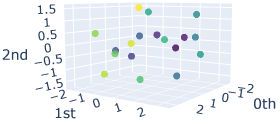}&
   \includegraphics[height=5.5cm]{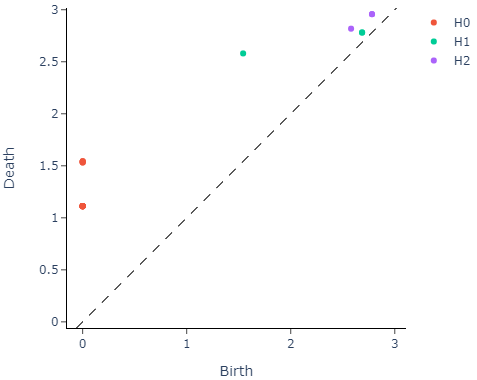}\\   
    \hline
  \end{tabular*}
   \caption{Different persistence diagrams for some C$_6$H$_x$ molecules. Persistent homology can capture the difference between the boat and chair isomers of cyclohexane.}
 \label{fig:6_C}
\end{figure*}

With the example of benzene in section~\ref{sec:benzene} we are now equipped to read the persistence diagrams. It is now apparent that a chain like hexane (See figure~\ref{fig:6_C}) will only have clusters, and the persistence plot clearly shows two points for the two bond length. 

Figure~\ref{fig:6_C} shows persistence diagrams for a set of $C_6 H_x$ molecules and here we can see how the low persistence features do allow us to differentiate between the different structures. The cyclohexane's persistence diagrams differ from benzene in two ways. Firstly the multiplicity of the $H_0$ features are different (but not visible on the plot). Secondly the cyclohexane breaks the symmetry of the loop, with the chair configuration having two voids: the 3-D holes above and below the plane, and the boat configuration gaining an extra loop from the top of the boat, and two voids. In chemistry it seems the less persistent features are still essential information. 

%!!! add in the actual numbers of the features of this later

%\subsection{Vectorising persistence diagrams}

%For example, for benzene we have

%[[table!]]

%persistence entropy for benzene [3.44756289, 0.        , 0.        ]

%[[put in the others?]]

%Note that this vector does not increase in size (as in length) of the input regardless of the size of the molecule.

%[[Put in benzene example here and a protein example]]

%Interesting fact: in the Molecule net paper it says that graph methods outperform standard methods 11/17 datasets. Have a look at how well you do...

Homology only counts topological shape, but the filtered complexes used in persistent homology include some aspect of geometry due to the fact that the simplical complexes are built from points in space and their relations. As a result, the two different forms of cyclohexane are demonstrably different, see their persistence diagrams in figure~\ref{fig:6_C}. Note that graph topology would not differentiate between these two structures as the graph connection between the atoms is the same in both cases.

Persistent homology only looks at shape and loses information about atom types, so information about enantomers is largely lost, for example, the persistence diagram of S- and D-alanine are the same to three decimal places (and the difference at that level is likely due to numerical calculations).   

\section{Results\label{sec:results}}

PHF features are smaller than the other 1-D input features, see table~\ref{tab:feature_comparison}.

\begin{table}[]
    \centering
    \begin{tabular}{crr}
    \hline
    Featuriser & Size of features & Size of MTR  \\
    \hline
 %   PHF        & 2.23MB        & 1,039,001     & 1.32$\pm$0.38 & 0.69$\pm$ 0.16 \\
    \textbf{PHF}        & \textbf{36}        & \textbf{1,039,001}      \\
    Grid        & 2052      & 3,055,001      \\
    ECFP        & 1024      & 2,027,001     \\
    MACCS & 166 & 1,169,001 \\
    RdKit & 208 & 1,211,001 \\
    \end{tabular}
    \caption{Feature size and multitask regressor model size.}
    \label{tab:feature_comparison}
\end{table}

\subsection{QM7}

The PHF approach works on the QM task (at least on the regression model featurisations) which is interesting as it is much smaller and does not include the Coulomb matrix energies or chemical features of CM-eigenvalues or rdkit. T-test for two independent samples for CM\_eig input and PHF and CM\_eig and PCA-PHF were both had a p-value $1\times10^{-5}$ indicating that the TDAF approach is not statistically significantly different from the Coulomb matrix eigenvalue approach (this is because the CM\_eig has a larger variance and in the 10 repeat run has a very large MAE.

%20 repates

\begin{figure}
    \centering
    \includegraphics[width=0.5\textwidth]{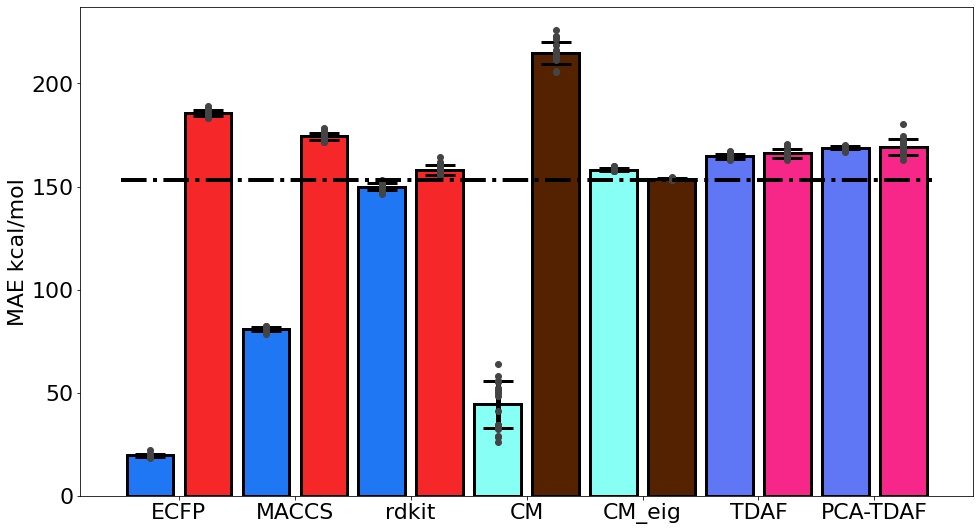}
    \caption{Singletask regression model for the QM7 dataset task (calculating atomisation energies). Blue, mint and purple are the results on the validation dataset for 1-D, Coulomb matrix based and PHF features respectively. Red, brown and pink are the results on the test dataset for 1-D, Coulomb matrix based and PHF features respectively.}
    \label{fig:QM7}
\end{figure}

%\subsection{QM8}

%Optimised hyperparameters. 

%Figure~\ref{fig:QM8_RMA} shows the results for QM8. The RF baseline indicates how difficult the problem is. Using topological features on a naive and un-hyperparameter-optimised NN gives the same quality result on [a?] test set as using ECFP with hyperparameter optimisation and training on a much larger dataset. Using hyperparameter optimisation and topological features [equals? beats? stats?] the state of the art on the test set, again using training with a smaller dataset. 

%So, topological features are information dense? Or have already done some of the processing? They make it possible to get the most out of small datasets (1/10th size of topol dataset). 

%\begin{figure}
%    \centering
%    \includegraphics[width=0.5\textwidth]%{2021_08_06_QM8_rmse_all_tests_run2.png}
%    \caption{RF: random forest; keras: kerasNN with topological features and no optimisation, ECFP was optimised and topol\_f was the topological features on an optimised NN. (Optimised means a hyperparameter search was done and the best model taken).Dot-dashed line is: ECFP was also trained on the whole dataset [[x]] molecules[[suspect theres an error here in the ectp numbers]}
%    \label{fig:QM8_RMSE}
%\end{figure}

\subsection{Lipophilicity}

\begin{figure}[htp]
    \centering
    \includegraphics[width=0.55\textwidth]{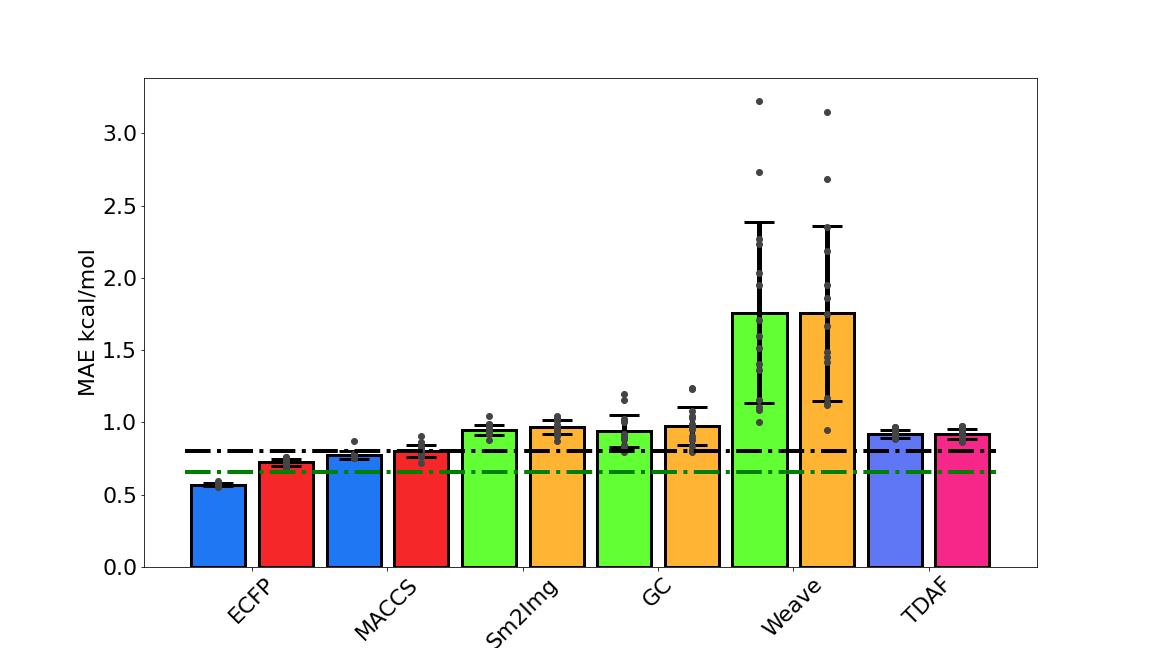}
    \caption{Performance on the lipophilicity dataset. Blue, green and purple are the results on the validation dataset for 1-D, 2-D and PHF features respectively. Red, orange and pink are the results on the test dataset for 1-D, 2-D and PHF features respectively.}
    \label{fig:lipo}
\end{figure}

The results for the lipophilicity dataset are given in figure~\ref{fig:lipo}. PHF works as well as the best 1-D approaches ECFP and as well as the best graph method (GC). There was no statistical difference between PHF and ECFP, MACCS and GC results. 

\subsection{Delaney (Solubility) dataset}

PHF works as well as the best featurisations on the Delaney dataset task. There was no statistically significant difference between PHF and the ECFP, MACCS and rdkit featurisations.

\begin{figure}[htp]
    \centering
    \includegraphics[width=0.55\textwidth]{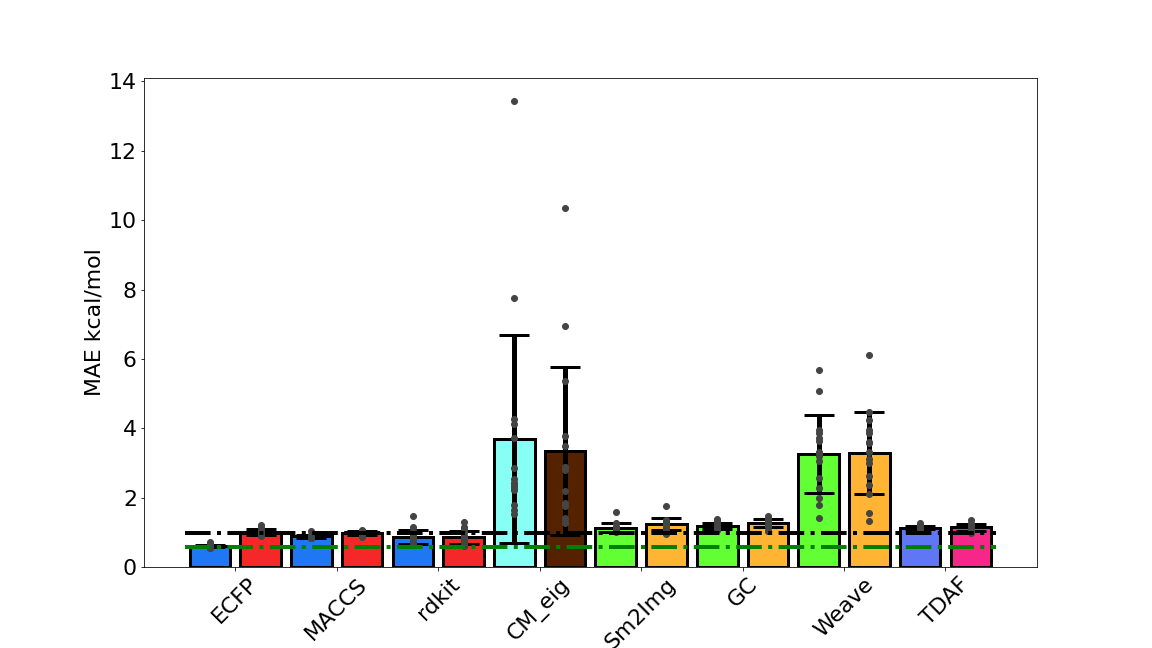}
    \caption{Performance on the Delaney dataset. Blue, mint, green and purple are the results on the validation dataset for 1-D, Coulomb matrix based, 2-D and PHF features respectively. Red, brown, orange and pink are the results on the test dataset for 1-D, Coulomb matrix based, 2-D and PHF features respectively.}
    \label{fig:delaney}
\end{figure}

\subsection{Sampl dataset}

The sample dataset is tiny (around 600 points) and here the ECFP and MACCS approaches clearly beat the PHF approach. 

\begin{figure}[htp]
    \centering
    \includegraphics[width=0.55\textwidth]{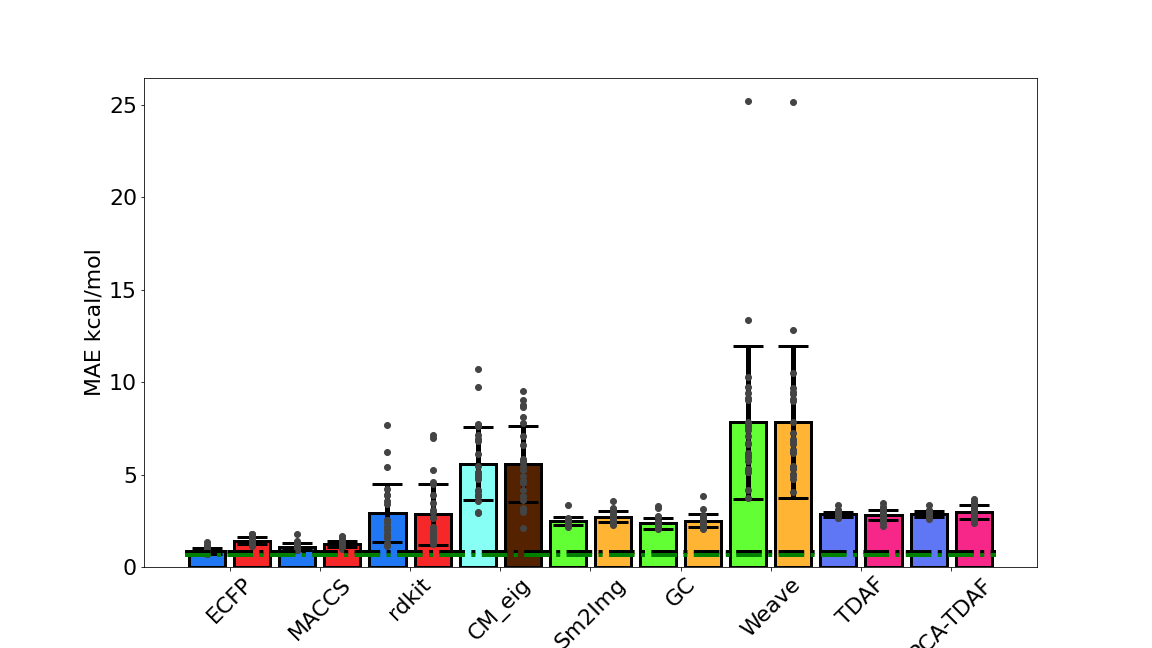}
    \caption{Performance on the Sampl dataset. Blue, mint, green and purple are the results on the validation dataset for 1-D, Coulomb matrix based, 2-D and PHF features respectively. Red, brown, orange and pink are the results on the test dataset for 1-D, Coulomb matrix based, 2-D and PHF features respectively.}
    \label{fig:sampl}
\end{figure}

%% !! to do check units for sample and delaney, check you're using the correct error

%\begin{table}[]
%%%% put p values in here???
%    \centering
%    \begin{tabular}{c|c}
%         &  \\
%         & 
%    \end{tabular}
%    \caption{Caption}
%    \label{tab:my_label}
%\end{table}
%lipo
%maccs
%a is 0.921512686746206 plus or minus 0.034468160019286896
%b is 0.8018943452856961 plus or minus 0.0428644479041223
%Ttest_indResult(statistic=-1.453983097931873, pvalue=0.15706920524570037)
%ecfp
%a is 0.921512686746206 plus or minus %0.034468160019286896
%b is 0.7256819433387797 plus or minus %0.02457309935843765
%Ttest_indResult(statistic=17.309680984417533, pvalue=1.7277203381802005e-16)
%gc
%a is 0.921512686746206 plus or minus 0.034468160019286896
%b is 0.9743141363803871 plus or minus 0.13143398518752242
%Ttest_indResult(statistic=-1.453983097931873, pvalue=0.15706920524570037)
%weave
%a is 0.921512686746206 plus or minus 0.034468160019286896
%b is 1.7545262143460323 plus or minus 0.6035631886726278
%Ttest_indResult(statistic=-5.155684081123022, pvalue=1.8120709011653895e-05)
%pca
%a is 0.921512686746206 plus or minus 0.034468160019286896
%b is 1.7661408741593692 plus or minus 0.8007818746782285
%Ttest_indResult(statistic=-5.155684081123022, pvalue=1.8120709011653895e-05)
%cm_eig
%a is 0.921512686746206 plus or minus 0.034468160019286896
%b is 4.64876280332274 plus or minus 3.320186945640462
%Ttest_indResult(statistic=-5.155684081123022, pvalue=1.8120709011653895e-05)

%\subsection{TDAF + rdkit features}

\section{Conclusions}

In this paper I have developed a novel molecular featurisation based on persistent homology called PHF (or TDAF for topological data analysis featurisation). This featurisation removes most of the chemically relevant symbols like atom type, bonds, valence, charge. Topological invariants like the number of connected regions, number of flat holes and voids are retained, and some small aspect of geometry is also encoded due to the use of filtration. Some chemical information can be inferred from these data, but the contents are largely topological shape invariants. I have shown that these PHFs can equal the state of the art on a series of chemical datasets like lipophilicity, solubility and QM7.

It is amazing that the PHFs work at all, given how little chemical information is encoded and this result should profoundly change our view of chemistry. I hesitate to say that 3-D shape is all that matters, but it is certainly more important that perhaps it was understood. There is however a proviso, neural networks do not `think' like us,\cite{GaleObjectDetectors, goodfellow2014explaining, 89} and neither should we be seduced into expecting them to (currently). As such, this result is not a suggestion that human chemists drop all their symbols and labels that we use to encode chemical information, merely that we be willing to let those symbols go when programming statistical learning machines, i.e. machine learning algorithms. 

The PHFs are very small data (18 number long vectors) and are thus smaller than any other chemical featurisations yet discovered. They are correspondingly very information dense. It should be pointed out that it does take longer to featurise molecules using this approach than with some of the simpler featurisers (like ECFP for example). This small dataset size means that we can use simpler and cheaper models (see table~\ref{tab:feature_comparison}), with lower energy usage, and a less of a requirement for huge, expensive computers, and this also helps to democratize machine learning away from cash-rich multinational companies towards students working with cheap laptops. 

The fact that PHFs work so well also proves that we can greatly improve machine learning by applying human knowledge and creativity to recast the problem into simpler terms (i.e. make the problem easier for the machine), and that this is a better approach than just throwing as much data and information at a complex model. The fact we can use smaller and simpler models reduces our requirement for fancy mathematics to prevent over-fitting and our need for huge datasets (although, as chemistry datasets are so small, we still need more data).

\section{Appendix: Code example\label{sec:code}}

Example code to calculate the persistence diagram of benzene. First we load our packages (note that using graphs-and-topology-for-chemists.py will do this for you.
\begin{verbatim}
from rdkit import Chem
import rdkit.Chem.AllChem as AllChem
from gtda.homology import VietorisRipsPersistence
from gtda.diagrams import PersistenceEntropy
import numpy as np
\end{verbatim}

This function creates coordinates from rdkits molecular structure. It is part of graphs-and-topology-for-chemists.py, but shown here in full fopr educational purposes.
\begin{verbatim}
def generate_structure_from_smiles(smiles):

    # Generate a 3-D structure from smiles

    mol = Chem.MolFromSmiles(smiles)
    mol = Chem.AddHs(mol)

    status = AllChem.EmbedMolecule(mol)
    status = AllChem.UFFOptimizeMolecule(mol)

    conformer = mol.GetConformer()
    coordinates = conformer.GetPositions()
    coordinates = np.array(coordinates)

    return coordinates 
\end{verbatim}

This function creates the Vietoris-Rips complex, calculates the persistence over length scales and creates the persistence diagram.
\begin{verbatim}
def smiles_to_persistence_diagrams(smiles):
    coords=generate_structure_from_smiles(smiles)
    # makes a point cloud version of the structure
    # there are no atom types

    # Track connected components, loops, and voids
    homology_dimensions = [0, 1, 2]

    # Peristence calculation
    persistence = VietorisRipsPersistence(
        metric="euclidean",
        homology_dimensions=homology_dimensions,
        n_jobs=6,
        collapse_edges=True,
    )
    reshaped_coords=coords[None, :, :]
    diagrams_basic=
        persistence.fit_transform(reshaped_coords)
    return coords, diagrams_basic
\end{verbatim}

\subsection{Creating the topological features\label{sec:code_example}}

This code block is all you need to type to go from a SMILES string to the small data input (just persistence entropy).
\begin{verbatim}
    smiles="c1ccccc1"
    coords, diagrams_basic
        =smiles_to_persistence_diagrams(smiles)
    persistence_entropy = PersistenceEntropy()
    # calculate topological feature matrix
    X_basic = 
        persistence_entropy.fit_transform(diagrams_basic)
\end{verbatim}

Code to produce the plots, do the larger number of features, and create an entire dataset of features are given in `graphs and topology for chemists'.

\section{Appendix: mathematical details\label{sec:maths}}

To learn more about topology, the interested reader is referred to Topology: illustrated, which nicely explains the mathematics with a huge number of pictures.\cite{saveliev2016topology}

\subsection{Algebraic topology details}

A \textit{space} is a set of mathematical objects that can be treated as points (although they may not be points) and selected relationships between them (rules for that space). For example, the the positions of atoms may be treated as points in a space.

\subsection{Metric space\label{ssec:metric_space}} \textit{A metric space} is a space of points (for example, $\{x,y,z,a,b\}$) defined by a function with the `rules' of a metric\footnote{A metric is literally a ruler}, $d$, \emph{distance function} which follows standard distance metric rules:

\begin{equation}
    d(x,y) = 0 \Leftrightarrow x = y \;,
    \label{eq:distance}
\end{equation}
i.e. if the distance between 2 points $x$ and $y$ is zero the points must be the same and \textit{vice versa},
\begin{equation}
    d(x,y) = d(y,x) \;,
\end{equation}
and the triangle inequality:
\begin{equation}
    d(x,z) \leq d(x,y) + d(y,z) \;.
\end{equation}

The Euclidean distance function in a 3-D space satisfies this and is a natural space to describe the 3-D molecular structure in. The Euclidean distance function is also known as the $l^2$ norm, as given by $x = \sqrt{(i^2 + j^2 + k^2)}$ in 3-D space where $x$ is a `point' in that space and $i$, $j$ and $k$ are the \emph{distances} along the Cartesian axes (not the unit vectors, sorry for abuse of terminology). Using the dot product and Euclidean norm you can create a vector space (literally a space where the `points' are vectors), $\mathbb{R}^3$, which is a vector space of 3-tuples of real numbers, the most familiar of which is Cartesian coordinates, $(i, j, k)$ and the vectors are defined from the origin $(0, 0, 0)$ (and this is what I referred to in this paper when talking about 3-D space).

A (3-D) point cloud is a finite subset of $\mathbb{R}^3$ with a metric induced from a Euclidean space. As such, a 3-D point cloud sits in a finite metric space, $(X,d)$ (let $d$ be the Euclidean distance function and $X$ is a list of points with some order: $X= x_1 < x_2 < x_3 ... < x_n$). From this we can make a distance matrix, where the $(x_i, x_j)$ entry in it is the distance between $x_i$ and $x_j$, $d(x_i, x_j)$. 

Thus, the atomic positions of the atoms in a molecule can be described as a point cloud with a Euclidean distance function as the metric. 

\subsection{Homology \label{ssec:homology}}

\begin{figure*}[htpb]
    \centering
    \includegraphics[width=7.5cm]{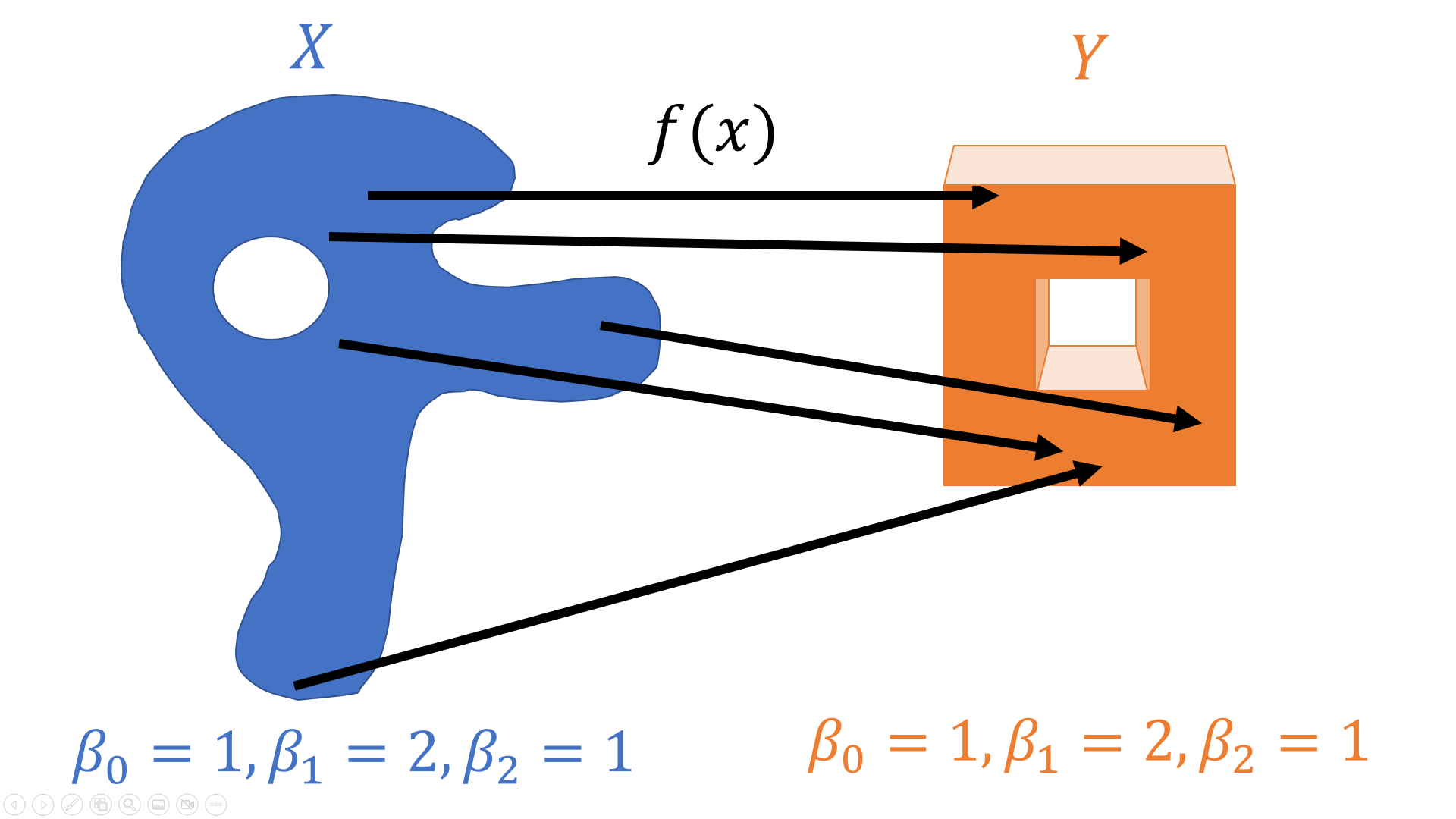}
    \includegraphics[width=8.5cm]{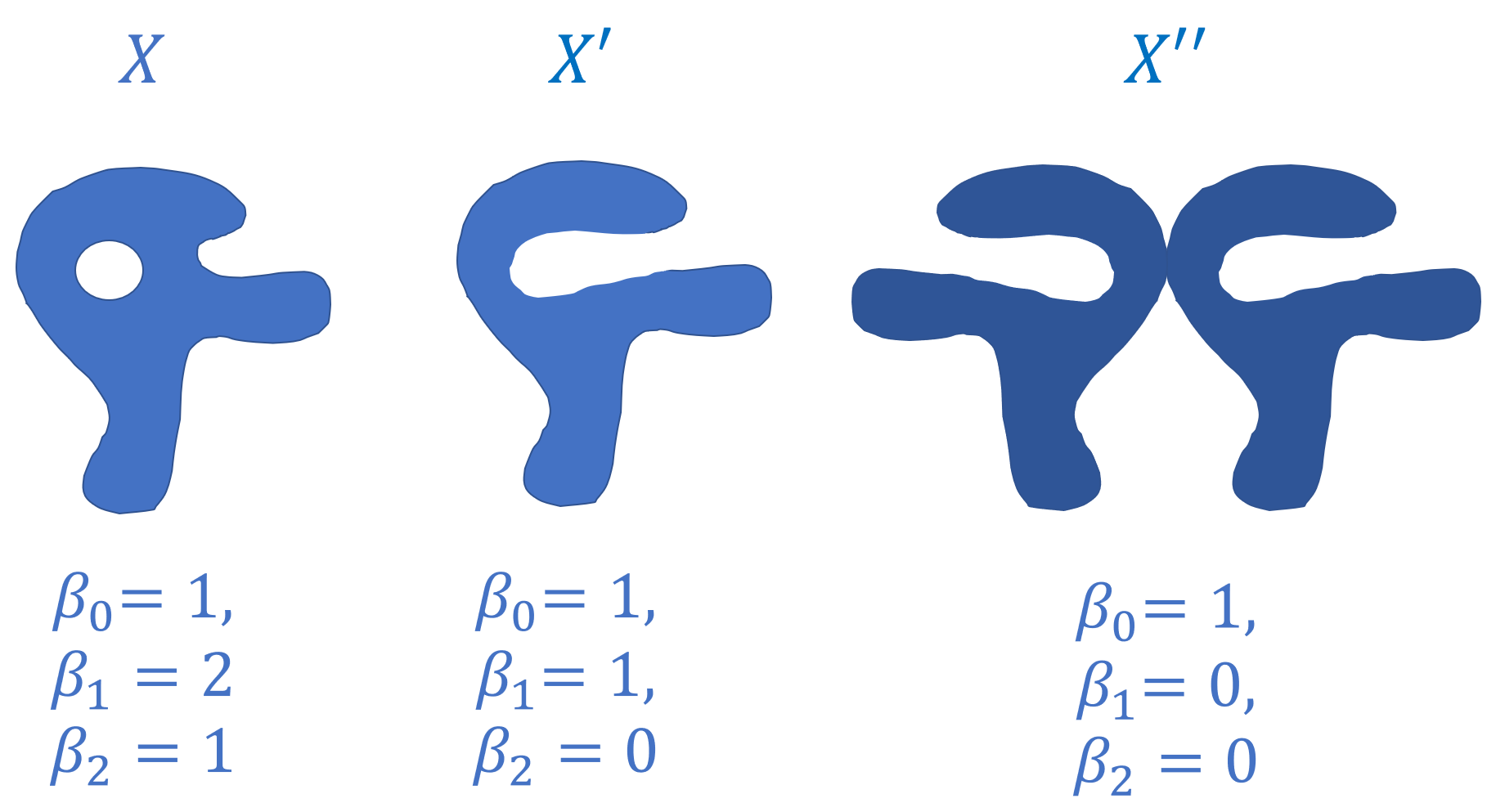}\\
    a. Continuous mapping from $X$ to $Y$.    b. Cutting $X$ changes its class but does not disconnect it.
    \includegraphics[width=7.5cm]{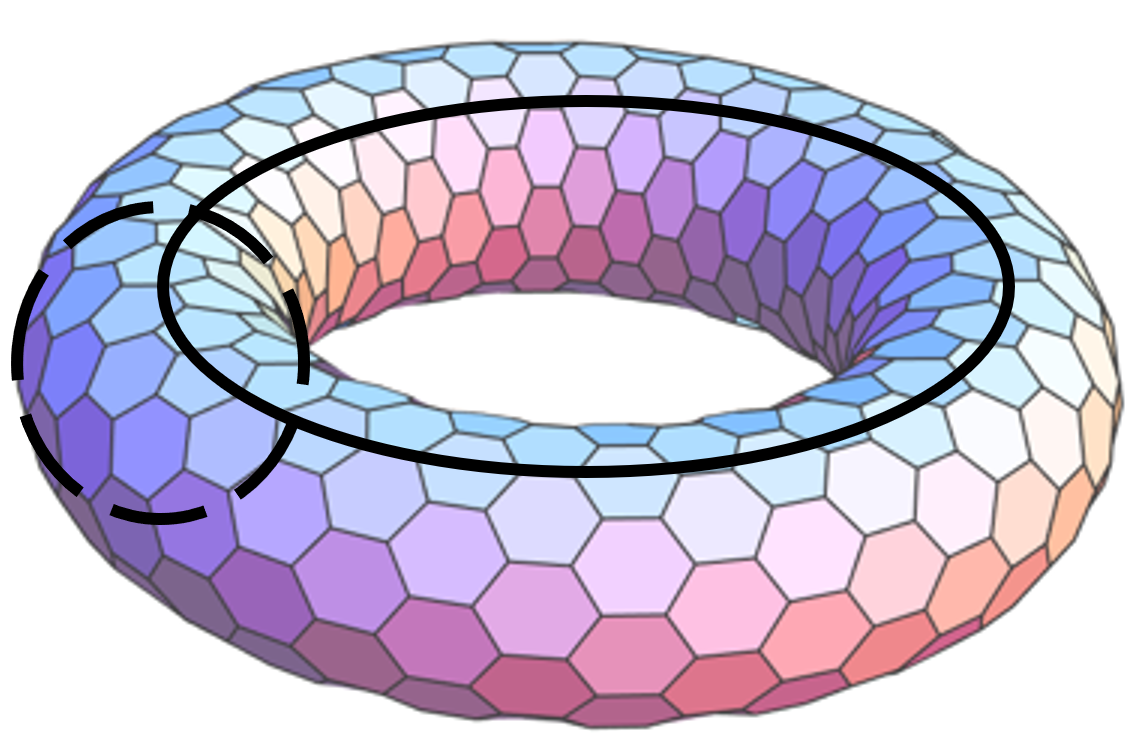}
    \includegraphics[width=7.5cm]{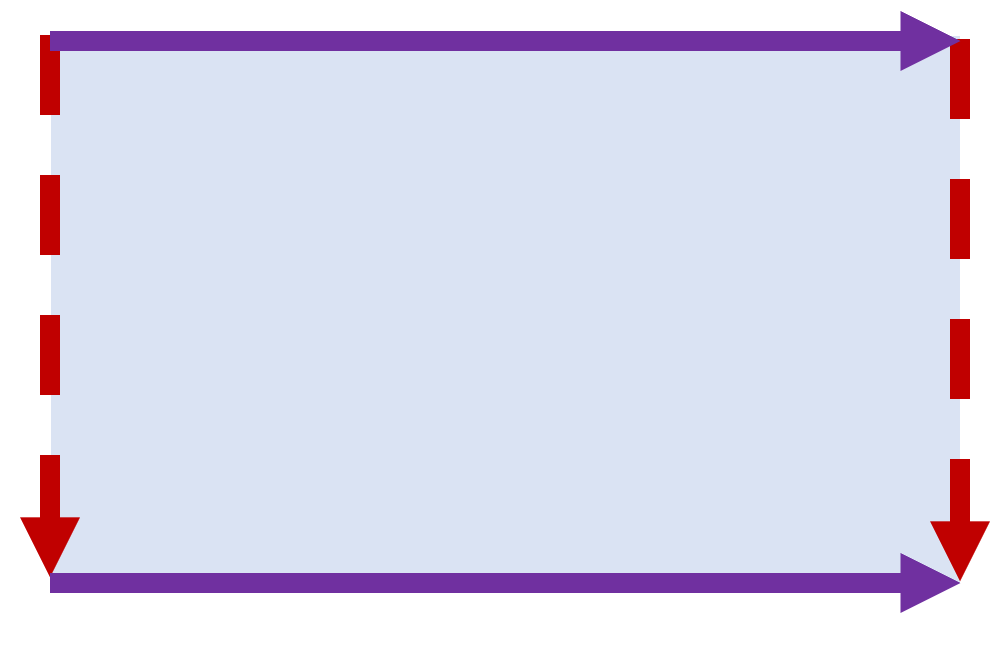}\\
    c. Generators on a torus d. Torus map
    \caption{Homological concepts. a. The 2-D topological spaces $X$ and $Y$ are homeomorphically equivalent to each other and a torus. Function $y=f(x)$ maps $X$ onto $Y$ via a deformation. Both the blue and orange regions are themselves connected, but disconnected from each other. b. Cutting the surface $X$ changes its class from torus to cylinder/circle and $X'$ is still connected. It requires two cuts to flatten the manifold, hence why $\beta_1 = 2$ in the original $X$. $X''$ is homeomorphically similar to a flat sheet. c. To find out the topology of a torus look at paths on the torus. There are an infinite number of paths that go around the ring, including those that loop around it more than once, but these are homotopically equivalent, and thus we have a generator $H_1$ which generates all these degenerate paths. There is a second $H_1$ generator which loops around the hole at the interiour of of the torus and generates an infinite set of paths. Thus $\beta_1 = 2$ in for a torus. The circles drawn on the torus are 1-cycles generated by the $H_1$ homology group. d. A torus can be made from a flat sheet by bending it into a cylinder and gluing the purple edges together so the arrows match, then bending the cylinder around to glue the dashed, red edges so that the arrows match.}
    \label{fig:math_concepts}
\end{figure*}

In $\mathbb{R}^3$ we can only have homology features up to dimension 2, i.e. our topological spaces are surfaces (manifolds of dimension 2). Figure~\ref{fig:math_concepts} presents an example of a torus surface~\ref{fig:math_concepts}c made from made from a stretchy sheet 2-D sheet~\ref{fig:math_concepts}d. The torus has two generators of dimension 2: $H_2$ and these generate families of curves one set that circle the large central hole, an example of one of these is drawn on the torus in black) one or more times and another set that circles the inside of the torus one or more times (an example of one of the drawn on the torus as a dashed black line. These circles are called 1-cycles in the language of topology, and all 1-cycles in the same group are homotopically equivalent to each other as they can be smoothly deformed into each other. There exists so-called trivial 1-cycles on the surface of the torus that do not encapsulate a hole, and these are homotopically equivalent to a point (as they can be transformed (i.e. shrunk)) to a point where they overlap. Note, that this is not a homeomorphism as one the 1-cycle has become a point, it cannot be reversed to recreate the 1-cycle. 

The manifold $X$ in figure~\ref{fig:math_concepts} exists in 3-D space. There is a continuous mapping $y=f(x)$ that can deform $X$ into $Y$ and both of these manifolds are homotopically (and homeomorphically, the reverse function $x=g(y)$ is not drawn on the figure) equivalent to the torus in figure~\ref{fig:math_concepts}c. 

If we cut through one of the generators in $X$ we create $X'$ which changes its topological class ($X$ and $X'$ are not homotopically equivalent) as the number of homotopical invariants has changed: specifically, the new manifold has one less hole, so $\beta_1$ is now 1. There is still the hole at the centre of the object, and $X'$ is now homotopically equivalent to a cylinder, and as a cylinder can be squashed to a circle, homotopically equivalent to a circle as well. The second operation is not homeomorphic for the same reason that contracting a circle to a point was. We can cut the remaining hole to make manifold $X''$ which has now been opened and is homeotopically equivalent to a flat sheet. As it took two cuts to flatten $X$, $\beta_1 = 2$, and this is also the number of generators, $H_2$, on the torus. 

%A \textit{manifold} is a space similar to Euclidean space. Homology is the study of mappings of manifolds. We expect the spaces to be connected and allow them to be deformed as if they were made of stretchy, bendy material (topological deformation). Any property that is conserved under this sort of operation is a topological invariant. Cutting, sticking and glueing types of operations are not topologically invariant. The invariants that are preserved are things like the number of holes of any dimension. A hole of dimension 0 is a connection, so the number of connected regions are also topological invariants. 

%The homology group $H_k(X)$ is a measure of the number of holes and connected regions of a space. The Betti numbers are related, as these are the rank of the homology groups and the number of cuts you can make in the object before it would be cut into separated regions. For example, a ring opening reaction cuts a molecule nd changes its homology group, but does not stop the molecule from being connected. A hole is something that is not there, so it is hard to think about how you might measure it, and the solution is based on paths on the surface. 

%Persistent homology is the extension of these ideas to point clouds.

%connected regions and number of holes (of different dimension). 

\subsection{Simplices\label{sec:simplices}}

\subsubsection{Flatness\label{ssec:flatness}} Now we will try to make the concept of `flatness' more rigorous.

Given a set of $n$ points $\{a_1, a_2, ... a_n \}$ a convex combination of these points is any point $x$ given by
\begin{equation}
    x = \sum_{i=0}^{n} \lambda_i a_i \;,
    \label{eq:constraints}
\end{equation}
%% !!! I'm not sure that that equation is correct, check it before publication
where $\lambda$ is real and must satisfy:
\begin{enumerate}
    \item $0 \leq \lambda_i \leq \lambda 1 ;, \forall i$
    \item $\sum_{i} \lambda = 1$.
\end{enumerate}
The \emph{convex hull} is the set of all convex combinations (i.e. the set of all points that satisfy the constraints).

Imagine a two points $\{a, b\}$ in $\mathbb{R}^3$, which are vectors from the origin and which are not parallel to any of the Cartesian axes. The first constraint keeps the points in the convex hull between $a$ and $b$, the second keeps them on the plane. The linear combinations are those that have no constraints on $\lambda_i$ in equation \ref{eq:constraints}, and this is the plane defined by $(0, a, b)$, and this can go to 2-D $\pm \infty$. An affine hull only has the second constraint, and this defines a line that sits in the plane $(0, a, b)$ and runs through the points $(a, b)$ and extends beyond them to $+\infty$ and $-\infty$ (i.e. 1-D). The convex hull is the line that sits in the plane $(0, a, b)$ and joins $a$ to $b$ and is thus finite, i.e. the line you would draw between points $a$ and $b$. It is the shortest straight line between those points and is thus `flat'. The convex hull of a triangle is the 3 straight lines between points $(a,b,c)$ and these of course sit in the plane $(a, b, c)$. The points in the convex set must also be geometrically independent, i.e. none are the same as another in the set.
%[[put in a picture ffs]]

\subsubsection{Convex sets and simplical complexes\label{ssec:simplical_complex}}

Convex sets (of which simplices are an example) are \textit{path-connected} (i.e. you could walk across the set) and have no holes or voids etc, instead it surrounds the holes and voids and this is one way that topology finds these features. A simplical complex, $X$, is a superset of all simplices found. A \emph{simplicial complex}  is a finite union of simplices in $\mathbb{R}^3$ satisfying that every face of a simplex in  is in  and that the non-empty intersection of two simplices in  is a face of each.

A \emph{abstract simplical complex} is a pair of sets $(V,X)$ with the elements of $X$ being subsets of V such that for every $v$ (vector) in $V$ (vector space), the singleton ${v}$ is also in $X$ and if $x$ is in $X$ and $y$ is a subset of $x$, then $y$ is in $X$. This defines the hierarchical structure of simplical complexes), see figure~\ref{fig:VR_complex}

The elements of $X$ are called simplices and the dimension of a simplex $x$ is defined by $|x| = \#x -1$ where $\#x$ denotes the cardinality of $x$. Simplices of dimension $n$ are called $n$-simplices. This is why tetrahedron is made of 4 triangles but is simplex-3, a triangle made of 3 lines is simplex-2 and a line made up of 2 points is simplex-1. Note although we keep to the simplices, you can use other shapes, for example a set based on hypercubes is a cubical complex (point, line, square, cube, hypercube), but this includes more simplices at each level (1, 2, 4, 8) than simplical complexes (1,2,3,4) so is less efficient; cell complexes are the other type of complex which use points, closed segments, disks, balls, hyperballs.

A \textit{filtered complex} is a collection of simplical complexes $\{X_\eta\}_{\eta \in \mathbb{R}^3}$ such that $X_\eta$ is a subcomplex of $X_t$ for each $\eta\leq t$. So filtered complexes are built up form simpler complexes and complexes with a higher value of $\eta$ contain those build with a lower value of $\eta$. 

%where $\eta$ is the radius of the ball and t is the radius of the maximum size of ball when everything dies. [[change other s's to eta.]

A Veitoris-Rips complex is a filtered complex built using a distance function. Let $(X,d)$ be a finite metric space (i.e. Euclidean distance metric space of a molecular point cloud). The Vietoris-Rips complex of $X$ is the filtered complex $V R_\eta(X)$ that contains a subset of $X$ as a simplex if all pairwise distances in the subset are less than or equal to $\eta$. The definition for a Vietoris-Rips complex is

\begin{equation}
    V R_{\eta(X)} = \{ [v_0, ... v_n | \forall i,j, d(v_i, v_j) \leq \eta \} \;.
\end{equation}
Basically, a Veirtoris-Rips complex is a set of simplical complexes that are formed by joining simplices if they are less than $\eta$ apart, it includes the sub-simplical complexes that are formed for values less than $\eta$. Figure ~\ref{fig:benzene_simplex} shows the simplical complexes for different values of $\eta$.

%multiplicity of point $(a_i, a_j)$
 %is the alternative sum of persistent% Bettig numbers
%V is a vector space

\paragraph{Homeomorphic} From the Greek `homoios' (similar) and `morpho' (shape) and means that the shape of two objects are similar because they are topologically equivalent. 

If $X$ and $Y$ are topological spaces and $f:X \rightarrow Y$ is a function that maps X onto Y, this function is homeomorphic if $f$: 

$$f(x) = f(y) \implies x = y \;,$$

for every $y \in Y$ there is an $x \in X$ such that $y = f(x)$ i.e. $x$ maps onto $y$, $f$ is continuous (no gaps) and $f^{-1}$ is continuous (i.e. there is an inverse and it is also has no gaps).\cite{saveliev2016topology} This means we can deform $X$ into $Y$ and back. 

The properties that remain constant during this deformation are \textit{topological invariants} and these are the properties that we count in this field, i.e. connected components, holes, voids etc. The properties which do not remain constant are those we lose in topology, i.e. distances, angles, geometry, chirality, volume etc.

\paragraph{Homotopy and homotopy equivalence}

If $f$ and $g$ are maps, $X$ and $Y$ are spaces,
%[[from topology and data]]
two continuous maps, $f, g : X \rightarrow Y$ are homotopic if there is a continuous map $H : X \times [0,1]$ such that $H(x,0)=f(x)$ and $H(x,1) = g(x)$, i.e. $H$ maps from $f$ to $g$ (and \textit{vice versa}) via a smooth deformation, i.e. we could mould a stretchy, material from one shape to another.

Two spaces $X$ and $Y$ are homotopy equivalent if there exists a map $G : Y \rightarrow X$ for a map $f : X \rightarrow Y$ so that $f \cdot y$ is homotopic to the identity map on $Y$ and $g \cdot f$ is homotopic to $f$.

For any topological space $X$ and Abelian group $A$ and integer $k$, $k > 0$, there is an assigned \textit{group} $H_k(X,A)$. If $f$ and $g$ are homotopic, then $H_k(f,A) = H_k(g,A)$. If $X$ and $Y$ are homotopy equivalent then $H_k(X,A)$ is isomorphic to $H_k(Y, A)$.

\paragraph{Betti numbers \label{ssec:betti}}
For any field, $F$, $H_k(X,F)$ will be a vector space over $F$. its dimension (if finite dimensional) is the Betti number $\beta_k(X,F)$ (And has coefficients in F). The $k$th Betti number is the number of independent $k$-dimensional surfaces. If two spaces are homotopy equivalent then all their Betti numbers are equal. (e.g. the surface of a donut and the surface of a tea-cup are homotopy equivalent as their Betti numbers are: $\beta_0 = 1$, $\beta_1=2$ and $\beta_2=1$. %(Betti numbers can vary with field, but I htink we are only using 3-D space.

%Betti numbers are the ranks of each group (the nested simplical compexes are Albelian groups) i.e. how many points are in them,
%'. Persistent homology is one of the mostused techniques for computing the topological invariants of a topological space. Itreturns a parametrized version of the Betti numbers: the Betti barcodes (see for example Fig. 11.1) [5]. The barcodes are equipped with the generators of the topologicalfeature (connected components, holes, voids, etc.). Generators are the set of nestedsimplices forming the topological features.'

\subsection{Multiset\label{ssec:mulitset}}

A mathematical \emph{set} is a set of \emph{elements} with no defined order over them, for example: $\{1, 5, 3, 2\}$ or $\{a, b, c, d\}$, where $\{ \}$ is commonly used to denote a set. We humans might infer a numerical or alphabetical ordering here, but there is no mathematical ordering when used. Items in a set are called \emph{elements}. A \emph{multiset} is the extension of this idea to allow multiple copies of an item, where the \emph{multiplicity} is the number of those items in the set, for example: $\{a, a, a, b, b, c\}$ is a multiset with element $a$ having a multiplicity of 3, element $b$ having a multiplicity of 2 and element $c$ having a multiplicity of 1. For example, a persistence diagram might have the multiset: $\{(b_1, d_1), (b_1, d_1), (b_1, d_1), (b_2, d_2), (b_3, d_3) \}$, where $(b_1, d_1)$ has a multiplicity of 3, $(b_2, d_2)$ and $(b_3, d_3)$ both have a multiplicity of 1.

\subsubsection{Persistence diagram\label{ssec:persistence_diagram}}

%Each topological feature is described by a pair of numbers, $\eta$, $t$ where $\eta$ is the current ball radius and $t$ is the maximum radius achieved. This is converted to a multiplicity of points $(b, d)$ where $b$ is the radius where the feature is born and $d$ is the radius where is dies. When plotted on a persistence diagram in 2-D the birth death 

A persistence diagram is a multiset of points containing the birth and death radii of all points with $d_i < +\infty$, i.e. we ignore the feature that continues to infinity. A persistence diagram, $D$, is written as $D = \{(b_i, d_i) \}_{i \in I}$

\subsection{Persistence entropy\label{ssec:persistence_entropy}}

The entropy of a persistence diagram, $E(D)$, is given by 

$$
E(D) = - \sum_{i \in I} p_i \log(p_i) \;,
$$

where $p_i$ is given by
$$
p_i \frac{(d_i - b_i) }{\sum_{i \in  I} (d_i - b_i ) } \;.
$$
As $(d_i - b_i)$ is the length of the barcodes it is the persistence of the feature, and persistence entropy measures the distribution in lengths of the barcodes. 

%\subsubsection{Wasserstein distance}

%kerber2017geometry - comp paper on how to do it

%%%% add this back in
%\section*{Author Contributions}
%We strongly encourage authors to include author contributions and recommend using \href{https://casrai.org/credit/}{CRediT} for standardised contribution descriptions. Please refer to our general \href{https://www.rsc.org/journals-books-databases/journal-authors-reviewers/author-responsibilities/}{author guidelines} for more information about authorship.
%%%% add this back in
%\section*{Conflicts of interest}
%In accordance with our policy on \href{https://www.rsc.org/journals-books-databases/journal-authors-reviewers/author-responsibilities/#code-of-conduct}{Conflicts of interest} please ensure that a conflicts of interest statement is included in your manuscript here.  Please note that this statement is required for all submitted manuscripts.  If no conflicts exist, please state that ``There are no conflicts to declare''.

\section*{Acknowledgements}
E.G. would like to thank EPSRC which funding this work, 
% grant code!
Oliver Matthews for discussions and proof-reading, the team that work on DeepChem and Peter Saveliev who published Topology Illustrated which helped make all these concepts easier.

%%%END OF MAIN TEXT%%%


\balance


%%%REFERENCES%%%
\bibliography{references, bristol} %You need to replace "rsc" on this line with the name of your .bib file

\providecommand*{\mcitethebibliography}{\thebibliography}
\csname @ifundefined\endcsname{endmcitethebibliography}
{\let\endmcitethebibliography\endthebibliography}{}
\begin{mcitethebibliography}{47}
\providecommand*{\natexlab}[1]{#1}
\providecommand*{\mciteSetBstSublistMode}[1]{}
\providecommand*{\mciteSetBstMaxWidthForm}[2]{}
\providecommand*{\mciteBstWouldAddEndPuncttrue}
  {\def\EndOfBibitem{\unskip.}}
\providecommand*{\mciteBstWouldAddEndPunctfalse}
  {\let\EndOfBibitem\relax}
\providecommand*{\mciteSetBstMidEndSepPunct}[3]{}
\providecommand*{\mciteSetBstSublistLabelBeginEnd}[3]{}
\providecommand*{\EndOfBibitem}{}
\mciteSetBstSublistMode{f}
\mciteSetBstMaxWidthForm{subitem}
{(\emph{\alph{mcitesubitemcount}})}
\mciteSetBstSublistLabelBeginEnd{\mcitemaxwidthsubitemform\space}
{\relax}{\relax}

\bibitem[Brown(2002)]{brown2002topology}
I.~D. Brown, \emph{Structural Chemistry}, 2002, \textbf{13}, 339--355\relax
\mciteBstWouldAddEndPuncttrue
\mciteSetBstMidEndSepPunct{\mcitedefaultmidpunct}
{\mcitedefaultendpunct}{\mcitedefaultseppunct}\relax
\EndOfBibitem
\bibitem[Coulson(1955)]{coulson1955contributions}
C.~A. Coulson, \emph{Journal of the Chemical Society (Resumed)}, 1955,
  2069--2084\relax
\mciteBstWouldAddEndPuncttrue
\mciteSetBstMidEndSepPunct{\mcitedefaultmidpunct}
{\mcitedefaultendpunct}{\mcitedefaultseppunct}\relax
\EndOfBibitem
\bibitem[Weisberg(2008)]{weisberg2008challenges}
M.~Weisberg, \emph{Philosophy of Science}, 2008, \textbf{75}, 932--946\relax
\mciteBstWouldAddEndPuncttrue
\mciteSetBstMidEndSepPunct{\mcitedefaultmidpunct}
{\mcitedefaultendpunct}{\mcitedefaultseppunct}\relax
\EndOfBibitem
\bibitem[Mezey(2012)]{mezey2012molecular}
P.~G. Mezey, \emph{Journal of Mathematical Chemistry}, 2012, \textbf{50},
  926--933\relax
\mciteBstWouldAddEndPuncttrue
\mciteSetBstMidEndSepPunct{\mcitedefaultmidpunct}
{\mcitedefaultendpunct}{\mcitedefaultseppunct}\relax
\EndOfBibitem
\bibitem[Gluck(1965)]{gluck1965chemical}
D.~Gluck, \emph{Journal of Chemical Documentation}, 1965, \textbf{5},
  43--51\relax
\mciteBstWouldAddEndPuncttrue
\mciteSetBstMidEndSepPunct{\mcitedefaultmidpunct}
{\mcitedefaultendpunct}{\mcitedefaultseppunct}\relax
\EndOfBibitem
\bibitem[Rogers and Hahn(2010)]{rogers2010extended}
D.~Rogers and M.~Hahn, \emph{Journal of chemical information and modeling},
  2010, \textbf{50}, 742--754\relax
\mciteBstWouldAddEndPuncttrue
\mciteSetBstMidEndSepPunct{\mcitedefaultmidpunct}
{\mcitedefaultendpunct}{\mcitedefaultseppunct}\relax
\EndOfBibitem
\bibitem[Lowis(1998)]{lowis1998hqsar}
D.~Lowis, \emph{Notes}, 1998, \textbf{1}, \relax
\mciteBstWouldAddEndPuncttrue
\mciteSetBstMidEndSepPunct{\mcitedefaultmidpunct}
{\mcitedefaultendpunct}{\mcitedefaultseppunct}\relax
\EndOfBibitem
\bibitem[Durant \emph{et~al.}(2002)Durant, Leland, Henry, and
  Nourse]{durant2002reoptimization}
J.~L. Durant, B.~A. Leland, D.~R. Henry and J.~G. Nourse, \emph{Journal of
  chemical information and computer sciences}, 2002, \textbf{42},
  1273--1280\relax
\mciteBstWouldAddEndPuncttrue
\mciteSetBstMidEndSepPunct{\mcitedefaultmidpunct}
{\mcitedefaultendpunct}{\mcitedefaultseppunct}\relax
\EndOfBibitem
\bibitem[Murray \emph{et~al.}(2016)Murray, Bellany, Benhamou, Bu{\v{c}}ar,
  Tabor, and Sheppard]{murray2016application}
P.~M. Murray, F.~Bellany, L.~Benhamou, D.-K. Bu{\v{c}}ar, A.~B. Tabor and T.~D.
  Sheppard, \emph{Organic \& biomolecular chemistry}, 2016, \textbf{14},
  2373--2384\relax
\mciteBstWouldAddEndPuncttrue
\mciteSetBstMidEndSepPunct{\mcitedefaultmidpunct}
{\mcitedefaultendpunct}{\mcitedefaultseppunct}\relax
\EndOfBibitem
\bibitem[Moseley and Murray(2014)]{moseley2014ligand}
J.~D. Moseley and P.~M. Murray, \emph{Journal of Chemical Technology \&
  Biotechnology}, 2014, \textbf{89}, 623--632\relax
\mciteBstWouldAddEndPuncttrue
\mciteSetBstMidEndSepPunct{\mcitedefaultmidpunct}
{\mcitedefaultendpunct}{\mcitedefaultseppunct}\relax
\EndOfBibitem
\bibitem[Durand and Fey(2019)]{durand2019computational}
D.~J. Durand and N.~Fey, \emph{Chemical reviews}, 2019, \textbf{119},
  6561--6594\relax
\mciteBstWouldAddEndPuncttrue
\mciteSetBstMidEndSepPunct{\mcitedefaultmidpunct}
{\mcitedefaultendpunct}{\mcitedefaultseppunct}\relax
\EndOfBibitem
\bibitem[Krizhevsky \emph{et~al.}(2012)Krizhevsky, Sutskever, and
  Hinton]{AlexNet}
A.~Krizhevsky, I.~Sutskever and G.~E. Hinton, Advances in neural information
  processing systems, 2012, pp. 1097--1105\relax
\mciteBstWouldAddEndPuncttrue
\mciteSetBstMidEndSepPunct{\mcitedefaultmidpunct}
{\mcitedefaultendpunct}{\mcitedefaultseppunct}\relax
\EndOfBibitem
\bibitem[Goodfellow \emph{et~al.}(2014)Goodfellow, Pouget-Abadie, Mirza, Xu,
  Warde-Farley, Ozair, Courville, and Bengio]{65}
I.~Goodfellow, J.~Pouget-Abadie, M.~Mirza, B.~Xu, D.~Warde-Farley, S.~Ozair,
  A.~Courville and Y.~Bengio, Advances in neural information processing
  systems, 2014, pp. 2672--2680\relax
\mciteBstWouldAddEndPuncttrue
\mciteSetBstMidEndSepPunct{\mcitedefaultmidpunct}
{\mcitedefaultendpunct}{\mcitedefaultseppunct}\relax
\EndOfBibitem
\bibitem[Szegedy \emph{et~al.}(2016)Szegedy, Vanhoucke, Ioffe, Shlens, and
  Wojna]{47}
C.~Szegedy, V.~Vanhoucke, S.~Ioffe, J.~Shlens and Z.~Wojna, Proceedings of the
  IEEE Conference on Computer Vision and Pattern Recognition, 2016, pp.
  2818--2826\relax
\mciteBstWouldAddEndPuncttrue
\mciteSetBstMidEndSepPunct{\mcitedefaultmidpunct}
{\mcitedefaultendpunct}{\mcitedefaultseppunct}\relax
\EndOfBibitem
\bibitem[Schwaller \emph{et~al.}(2020)Schwaller, Petraglia, Zullo, Nair,
  Haeuselmann, Pisoni, Bekas, Iuliano, and Laino]{schwaller2020predicting}
P.~Schwaller, R.~Petraglia, V.~Zullo, V.~H. Nair, R.~A. Haeuselmann, R.~Pisoni,
  C.~Bekas, A.~Iuliano and T.~Laino, \emph{Chemical science}, 2020,
  \textbf{11}, 3316--3325\relax
\mciteBstWouldAddEndPuncttrue
\mciteSetBstMidEndSepPunct{\mcitedefaultmidpunct}
{\mcitedefaultendpunct}{\mcitedefaultseppunct}\relax
\EndOfBibitem
\bibitem[Ball(2021)]{balllang}
P.~Ball, \emph{Chemistry World}, 2021\relax
\mciteBstWouldAddEndPuncttrue
\mciteSetBstMidEndSepPunct{\mcitedefaultmidpunct}
{\mcitedefaultendpunct}{\mcitedefaultseppunct}\relax
\EndOfBibitem
\bibitem[Gordin(2015)]{gordinlang}
M.~Gordin, \emph{Chemistry World}, 2015\relax
\mciteBstWouldAddEndPuncttrue
\mciteSetBstMidEndSepPunct{\mcitedefaultmidpunct}
{\mcitedefaultendpunct}{\mcitedefaultseppunct}\relax
\EndOfBibitem
\bibitem[Francl(2009)]{francl2009stretching}
M.~Francl, \emph{Nature chemistry}, 2009, \textbf{1}, 334--335\relax
\mciteBstWouldAddEndPuncttrue
\mciteSetBstMidEndSepPunct{\mcitedefaultmidpunct}
{\mcitedefaultendpunct}{\mcitedefaultseppunct}\relax
\EndOfBibitem
\bibitem[King(1991)]{king1991topological}
R.~King, \emph{Journal of Mathematical Chemistry}, 1991, \textbf{7},
  51--68\relax
\mciteBstWouldAddEndPuncttrue
\mciteSetBstMidEndSepPunct{\mcitedefaultmidpunct}
{\mcitedefaultendpunct}{\mcitedefaultseppunct}\relax
\EndOfBibitem
\bibitem[Fielden \emph{et~al.}(2017)Fielden, Leigh, and
  Woltering]{fielden2017molecular}
S.~D. Fielden, D.~A. Leigh and S.~L. Woltering, \emph{Angewandte Chemie
  International Edition}, 2017, \textbf{56}, 11166--11194\relax
\mciteBstWouldAddEndPuncttrue
\mciteSetBstMidEndSepPunct{\mcitedefaultmidpunct}
{\mcitedefaultendpunct}{\mcitedefaultseppunct}\relax
\EndOfBibitem
\bibitem[Sumners(1987)]{sumners1987knot}
D.~Sumners, \emph{Journal of mathematical chemistry}, 1987, \textbf{1},
  1--14\relax
\mciteBstWouldAddEndPuncttrue
\mciteSetBstMidEndSepPunct{\mcitedefaultmidpunct}
{\mcitedefaultendpunct}{\mcitedefaultseppunct}\relax
\EndOfBibitem
\bibitem[Adams \emph{et~al.}(2020)Adams, Devadoss, Elhamdadi, and
  Mashaghi]{adams2020knot}
C.~Adams, J.~Devadoss, M.~Elhamdadi and A.~Mashaghi, \emph{Journal of
  Mathematical Chemistry}, 2020, \textbf{58}, 1711--1736\relax
\mciteBstWouldAddEndPuncttrue
\mciteSetBstMidEndSepPunct{\mcitedefaultmidpunct}
{\mcitedefaultendpunct}{\mcitedefaultseppunct}\relax
\EndOfBibitem
\bibitem[Scalvini \emph{et~al.}(2021)Scalvini, Sheikhhassani, and
  Mashaghi]{scalvini2021topological}
B.~Scalvini, V.~Sheikhhassani and A.~Mashaghi, \emph{Physical Chemistry
  Chemical Physics}, 2021\relax
\mciteBstWouldAddEndPuncttrue
\mciteSetBstMidEndSepPunct{\mcitedefaultmidpunct}
{\mcitedefaultendpunct}{\mcitedefaultseppunct}\relax
\EndOfBibitem
\bibitem[Calladine and Drew(1997)]{calladine1997understanding}
C.~R. Calladine and H.~Drew, \emph{Understanding DNA: the molecule and how it
  works}, Academic press, 1997\relax
\mciteBstWouldAddEndPuncttrue
\mciteSetBstMidEndSepPunct{\mcitedefaultmidpunct}
{\mcitedefaultendpunct}{\mcitedefaultseppunct}\relax
\EndOfBibitem
\bibitem[Rouvray(1995)]{rouvray1995rationale}
D.~Rouvray, \emph{Journal of Molecular Structure: THEOCHEM}, 1995,
  \textbf{336}, 101--114\relax
\mciteBstWouldAddEndPuncttrue
\mciteSetBstMidEndSepPunct{\mcitedefaultmidpunct}
{\mcitedefaultendpunct}{\mcitedefaultseppunct}\relax
\EndOfBibitem
\bibitem[Fabi{\^c}-Petra{\^c} \emph{et~al.}(1991)Fabi{\^c}-Petra{\^c},
  Jerman-Bla{\v{z}}i{\^c}, and Batagelj]{fabic1991study}
I.~Fabi{\^c}-Petra{\^c}, B.~Jerman-Bla{\v{z}}i{\^c} and V.~Batagelj,
  \emph{Journal of mathematical chemistry}, 1991, \textbf{8}, 121--134\relax
\mciteBstWouldAddEndPuncttrue
\mciteSetBstMidEndSepPunct{\mcitedefaultmidpunct}
{\mcitedefaultendpunct}{\mcitedefaultseppunct}\relax
\EndOfBibitem
\bibitem[Basak \emph{et~al.}(1991)Basak, Niemi, and Veith]{basak1991predicting}
S.~C. Basak, G.~J. Niemi and G.~D. Veith, \emph{Journal of Mathematical
  Chemistry}, 1991, \textbf{7}, 243--272\relax
\mciteBstWouldAddEndPuncttrue
\mciteSetBstMidEndSepPunct{\mcitedefaultmidpunct}
{\mcitedefaultendpunct}{\mcitedefaultseppunct}\relax
\EndOfBibitem
\bibitem[Carlsson(2009)]{carlsson2009topology}
G.~Carlsson, \emph{Bulletin of the American Mathematical Society}, 2009,
  \textbf{46}, 255--308\relax
\mciteBstWouldAddEndPuncttrue
\mciteSetBstMidEndSepPunct{\mcitedefaultmidpunct}
{\mcitedefaultendpunct}{\mcitedefaultseppunct}\relax
\EndOfBibitem
\bibitem[Wasserman(2018)]{wasserman2018topological}
L.~Wasserman, \emph{Annual Review of Statistics and Its Application}, 2018,
  \textbf{5}, 501--532\relax
\mciteBstWouldAddEndPuncttrue
\mciteSetBstMidEndSepPunct{\mcitedefaultmidpunct}
{\mcitedefaultendpunct}{\mcitedefaultseppunct}\relax
\EndOfBibitem
\bibitem[Murugan and Robertson(2019)]{murugan2019introduction}
J.~Murugan and D.~Robertson, \emph{arXiv preprint arXiv:1904.11044}, 2019\relax
\mciteBstWouldAddEndPuncttrue
\mciteSetBstMidEndSepPunct{\mcitedefaultmidpunct}
{\mcitedefaultendpunct}{\mcitedefaultseppunct}\relax
\EndOfBibitem
\bibitem[Pirashvili \emph{et~al.}(2018)Pirashvili, Steinberg, Belchi~Guillamon,
  Niranjan, Frey, and Brodzki]{pirashvili2018improved}
M.~Pirashvili, L.~Steinberg, F.~Belchi~Guillamon, M.~Niranjan, J.~G. Frey and
  J.~Brodzki, \emph{Journal of cheminformatics}, 2018, \textbf{10}, 1--14\relax
\mciteBstWouldAddEndPuncttrue
\mciteSetBstMidEndSepPunct{\mcitedefaultmidpunct}
{\mcitedefaultendpunct}{\mcitedefaultseppunct}\relax
\EndOfBibitem
\bibitem[Delaney(2004)]{delaney2004esol}
J.~S. Delaney, \emph{Journal of chemical information and computer sciences},
  2004, \textbf{44}, 1000--1005\relax
\mciteBstWouldAddEndPuncttrue
\mciteSetBstMidEndSepPunct{\mcitedefaultmidpunct}
{\mcitedefaultendpunct}{\mcitedefaultseppunct}\relax
\EndOfBibitem
\bibitem[Atienza \emph{et~al.}(2016)Atienza, Gonzalez-Diaz, and
  Rucco]{atienza2016separating}
N.~Atienza, R.~Gonzalez-Diaz and M.~Rucco, Federation of International
  Conferences on Software Technologies: Applications and Foundations, 2016, pp.
  3--12\relax
\mciteBstWouldAddEndPuncttrue
\mciteSetBstMidEndSepPunct{\mcitedefaultmidpunct}
{\mcitedefaultendpunct}{\mcitedefaultseppunct}\relax
\EndOfBibitem
\bibitem[Saveliev(2016)]{saveliev2016topology}
P.~Saveliev, \emph{Topology illustrated}, 2016\relax
\mciteBstWouldAddEndPuncttrue
\mciteSetBstMidEndSepPunct{\mcitedefaultmidpunct}
{\mcitedefaultendpunct}{\mcitedefaultseppunct}\relax
\EndOfBibitem
\bibitem[Rucco \emph{et~al.}(2016)Rucco, Castiglione, Merelli, and
  Pettini]{rucco2016characterisation}
M.~Rucco, F.~Castiglione, E.~Merelli and M.~Pettini, Proceedings of ECCS 2014:
  European Conference on Complex Systems, 2016, pp. 117--128\relax
\mciteBstWouldAddEndPuncttrue
\mciteSetBstMidEndSepPunct{\mcitedefaultmidpunct}
{\mcitedefaultendpunct}{\mcitedefaultseppunct}\relax
\EndOfBibitem
\bibitem[Kerber \emph{et~al.}(2017)Kerber, Morozov, and
  Nigmetov]{kerber2017geometry}
M.~Kerber, D.~Morozov and A.~Nigmetov, \emph{Geometry helps to compare
  persistence diagrams}, 2017\relax
\mciteBstWouldAddEndPuncttrue
\mciteSetBstMidEndSepPunct{\mcitedefaultmidpunct}
{\mcitedefaultendpunct}{\mcitedefaultseppunct}\relax
\EndOfBibitem
\bibitem[Adams \emph{et~al.}(2017)Adams, Emerson, Kirby, Neville, Peterson,
  Shipman, Chepushtanova, Hanson, Motta, and Ziegelmeier]{adams2017persistence}
H.~Adams, T.~Emerson, M.~Kirby, R.~Neville, C.~Peterson, P.~Shipman,
  S.~Chepushtanova, E.~Hanson, F.~Motta and L.~Ziegelmeier, \emph{Journal of
  Machine Learning Research}, 2017, \textbf{18}, \relax
\mciteBstWouldAddEndPuncttrue
\mciteSetBstMidEndSepPunct{\mcitedefaultmidpunct}
{\mcitedefaultendpunct}{\mcitedefaultseppunct}\relax
\EndOfBibitem
\bibitem[Gale \emph{et~al.}(2018)Gale, Martin, and Bowers]{gale2018and}
E.~M. Gale, N.~Martin and J.~S. Bowers, \emph{arXiv preprint arXiv:1806.03934},
  2018\relax
\mciteBstWouldAddEndPuncttrue
\mciteSetBstMidEndSepPunct{\mcitedefaultmidpunct}
{\mcitedefaultendpunct}{\mcitedefaultseppunct}\relax
\EndOfBibitem
\bibitem[Wu \emph{et~al.}(2018)Wu, Ramsundar, Feinberg, Gomes, Geniesse, Pappu,
  Leswing, and Pande]{wu2018moleculenet}
Z.~Wu, B.~Ramsundar, E.~N. Feinberg, J.~Gomes, C.~Geniesse, A.~S. Pappu,
  K.~Leswing and V.~Pande, \emph{Chemical science}, 2018, \textbf{9},
  513--530\relax
\mciteBstWouldAddEndPuncttrue
\mciteSetBstMidEndSepPunct{\mcitedefaultmidpunct}
{\mcitedefaultendpunct}{\mcitedefaultseppunct}\relax
\EndOfBibitem
\bibitem[G.Landrum()]{rdkit}
G.Landrum, \emph{RDKit: Open-Source Cheminformatics Software},
  \url{http://www.rdkit.org/}\relax
\mciteBstWouldAddEndPuncttrue
\mciteSetBstMidEndSepPunct{\mcitedefaultmidpunct}
{\mcitedefaultendpunct}{\mcitedefaultseppunct}\relax
\EndOfBibitem
\bibitem[Tauzin(2020)]{Giotto-tda}
G.~Tauzin, \emph{Giotto-tda}, 2020,
  \url{https://github.com/giotto-ai/giotto-tda}\relax
\mciteBstWouldAddEndPuncttrue
\mciteSetBstMidEndSepPunct{\mcitedefaultmidpunct}
{\mcitedefaultendpunct}{\mcitedefaultseppunct}\relax
\EndOfBibitem
\bibitem[Tauzin \emph{et~al.}(2020)Tauzin, Lupo, Tunstall, P{\'e}rez, Caorsi,
  Medina-Mardones,\emph{et~al.}]{tauzin2020giotto}
G.~Tauzin, U.~Lupo, L.~Tunstall, J.~P{\'e}rez, M.~Caorsi, A.~Medina-Mardones
  \emph{et~al.}, \emph{arXiv preprint arXiv:2004.02551}, 2020\relax
\mciteBstWouldAddEndPuncttrue
\mciteSetBstMidEndSepPunct{\mcitedefaultmidpunct}
{\mcitedefaultendpunct}{\mcitedefaultseppunct}\relax
\EndOfBibitem
\bibitem[Gale(2022)]{GandT}
E.~M. Gale, \emph{GandT: Graphs and Topology for Chemistry}, 2022,
  \url{https://github.com/ellagale/GandT}\relax
\mciteBstWouldAddEndPuncttrue
\mciteSetBstMidEndSepPunct{\mcitedefaultmidpunct}
{\mcitedefaultendpunct}{\mcitedefaultseppunct}\relax
\EndOfBibitem
\bibitem[Gale(2020)]{DCval}
E.~M. Gale, \emph{DeepChem Validation}, 2020,
  \url{https://github.com/ellagale/deepchem\_2.3\_validation}\relax
\mciteBstWouldAddEndPuncttrue
\mciteSetBstMidEndSepPunct{\mcitedefaultmidpunct}
{\mcitedefaultendpunct}{\mcitedefaultseppunct}\relax
\EndOfBibitem
\bibitem[Gale \emph{et~al.}(2020)Gale, Martin, Blything, Nguyen, and
  Bowers]{GaleObjectDetectors}
E.~M. Gale, N.~Martin, R.~Blything, A.~Nguyen and J.~S. Bowers, \emph{Vision
  Research}, 2020, \textbf{176}, 60--71\relax
\mciteBstWouldAddEndPuncttrue
\mciteSetBstMidEndSepPunct{\mcitedefaultmidpunct}
{\mcitedefaultendpunct}{\mcitedefaultseppunct}\relax
\EndOfBibitem
\bibitem[Goodfellow \emph{et~al.}(2014)Goodfellow, Shlens, and
  Szegedy]{goodfellow2014explaining}
I.~J. Goodfellow, J.~Shlens and C.~Szegedy, \emph{arXiv preprint
  arXiv:1412.6572}, 2014\relax
\mciteBstWouldAddEndPuncttrue
\mciteSetBstMidEndSepPunct{\mcitedefaultmidpunct}
{\mcitedefaultendpunct}{\mcitedefaultseppunct}\relax
\EndOfBibitem
\bibitem[Marin~Dujmović(2020)]{89}
J.~B. Marin~Dujmović, Gaurav~Malhotra, \emph{bioRxiv 2020.02.25.964361; doi:
  https://doi.org/10.1101/2020.02.25.964361}, 2020\relax
\mciteBstWouldAddEndPuncttrue
\mciteSetBstMidEndSepPunct{\mcitedefaultmidpunct}
{\mcitedefaultendpunct}{\mcitedefaultseppunct}\relax
\EndOfBibitem
\end{mcitethebibliography}
\bibliographystyle{rsc} %the RSC's .bst file

\end{document}

\maketitle

\begin{abstract}

\end{abstract}

\section{Publicons for paper and supplementary information}

\section{Delaney Experiment}

measured log solubility in mols per litre

\subsection{$R^2$}

\begin{table}[]
    \centering
    \begin{tabular}{|c|c|}
        Name        & $R^2$ \\
        \hline
        Delaney Original\cite{delaney2004esol} & 0.69\\
         & \\
    \end{tabular}
    \caption{Caption .Units are log solubility in mols per litre.}
    \label{tab:Delaney_MAE}
\end{table}

\subsection{RMSE}

\subsection{Delaney MAE}

REsults below are the raw results, they've not been cleaned! (As in, we've not used the thingy that removes the outliers and adds a couple of percent to the answers).

\begin{table}[]
    \centering
    \begin{tabular}{|c|c|}
        name & R2 \\
        Random forest & \\
        NN? & \\
    \end{tabular}
    \caption{Pearson correlation coefficient for the PDBBind core dataset using a 80:10:10 train:validation:test split on a shuffled dataset, experiments were repeated 10 times (n=10), best value given in brackets. *score from the single best model [PDBBind v2016]}
    \label{tab:core_only}
\end{table}

\section{Experiment 4: Training on augmented and testing on core}

\begin{table}[]
    \centering
    \begin{tabular}{|c|c|c|c}
    \hline
        Name & trainable & no. & R$^2$ \\
            & parameters & desc.$^1$& \\
    \hline
    YAED-17f 3-head regression & ?& 17 & \textbf{0.92}\\
    yet another even deeper 3-head regression &  13mill & 15& \textbf{0.8997!}\\
    even deeper three-headed regression &  9,158,355  & 15 & \textbf{0.88}, 0.867$\pm ?$\\
    deep three headed regression & & 15 & 0.87 \\
    \textbf{deeper two-headed regression} & 181,425 & 0 & 0.85\\
    deep two-headed regression & 180,345 & 0 & 0.84\\
    two-headed regression model & & 0& ??\\
    \hline

    ES-IDM+ES-IEM GBT model\footnote{topological persistance features gradient based tree}\cite{meng2020persistent} & & & 0.840\\
    DeepAtom*\cite{rezaei2019improving} (augmented) &  & &  0.83\\
    K$_{deep}$*\cite{jimenez2018k} (augmented)[[cite orig paper]]  & 1,340,769 & & 0.82\\
    RF-Scorev3+RDKit ligand features$^{o}$\cite{boyles2020learning} & & & 0.821\\
    DeepAtom\cite{rezaei2019improving} &  & & 0.81\\
    RF-Score*\cite{rezaei2019improving} (augmented)[[cite orig paper]] & & & 0.81\\
    Mid-level Fusion model$^{+}$\cite{jones2020improved} & & 8& 0.810 \\
    ASFP \cite{zhang2020asfp} & & & 0.81 \\
    RF-Scorev3 features$^o$\cite{boyles2020learning} & & 185 & 0.814\\
    OnionNet \cite{zheng2019onionnet} & 143,882,777 & & 0.812($SD = 1.257$)\\
    RF-Score\cite{ballester2014does} & & & 0.803 \\
    3D-CNN (voxel) model$^{+}$\cite{jones2020improved} & & 8 or 21? & 0.723 \\
    Bappl+$^{\$}$\cite{soni2020improving} & & & 0.710\\
    DeepAtom\cite{rezaei2019improving} no hyperparameter refinement & & & 0.70\\
    SG-CNN model (graph)$^{+}$\cite{jones2020improved} & & 21 & 0.666 \\
    Best classical scoring function\cite{ballester2014does} X-Score::HMScore & & & 0.64 \\
    Mean classical scoring function\cite{ballester2014does} (n=18) & & & 0.427$\pm$0.03 \\
    Worst classical scoring function\cite{ballester2014does} SYBYL::F-Score & & & 0.22 \\
    \end{tabular}
    \caption{Pearson correlation coefficient (R$^2$) for test set using the standard protein binding training approach of train on the PDBBind refined dataset and test on the PDB core dataset and using with no validation. *score from the single best model.$^1$ Molecular descriptors refers to the number of molecular features input. In this count I do not count the coordinates, to include them increment the reported number for my work by 2 (protein coords and ligand coords). $^+$ this model was trained on both generalised and restricted dataset. $^{o}$ these models use the core dataset as both the valdiation and test sets. $^{\$}$ using RF to fit a MM forcefield [PDBBind v2016]}
    \label{tab:my_label}
\end{table}

\begin{table}[]
    \centering
    \begin{tabular}{|c|c|c|c}
    \hline
        Name &  R$^2$ \\
    \hline
    
    deeper two-headed regression & 0.81\\
    deep two-headed regression & \textbf{0.83}\\
    two-headed regression & ??\\    
    \hline
    DeepAtom*\cite{rezaei2019improving} (augmented)  & 0.79\\
    RF-Score*\cite{rezaei2019improving} (augmented)[[cite orig paper]]   & 0.75\\
    \end{tabular}
    \caption{Pearson correlation coefficient (R$^2$) for test set using the standard machine learning training approach of a randomly assigning complexes  80:10:10 train:validation:test datasets. The dataset used was a combination of the PDBBind refined and core datasets. *score from the single best model [PDBBind v2016]}
    \label{tab:my_label}
\end{table}

\begin{table}[]
    \centering
    \begin{tabular}{c|c|c|c}
    \hline
        Name & MAE \\
    \hline
    
    \hline 
    even deeper 3 headed regression & 0.539\\
    deep-3 headed regression & 0.588\\
    \hline
    OnionNet \cite{zheng2019onionnet} & 0.984 \\
     \cite{rezaei2019improving}  Autodock4[[rezaei2019improving:Morris2009]] & 2-3 kcal/mol\\
     ??[[rezaei2019improving:Gomes2017]]  & 1 kcal/mol\\
          Mid-level Fusion model$^{+}$\cite{jones2020improved} & 1.02 \\
     3D-CNN model$^{+}$\cite{jones2020improved} & 1.164 \\
     SG-CNN model$^{+}$\cite{jones2020improved} & 1.321 \\
     AK-score single\cite{kwon2020ak} & 1.511\\
     \hline
     AK-score ensemble\cite{kwon2020ak} & 1.101 \\ 
    \end{tabular}
    \caption{Caption}
    \label{tab:my_label}
\end{table}

\begin{table}[]
    \centering
    \begin{tabular}{|c|c|c|c}
    \hline
        Name & RMSE \\
    \hline
        even deeper 3 headed regression & 0.777\\
    deep-3 headed regression & 0.795\\
    \hline
    OnionNet \cite{zheng2019onionnet} & 1.278 \\
     DeepAtom*\cite{rezaei2019improving} (augmented)   & 1.1\\
     RF-Score*\cite{rezaei2019improving} (augmented)[[cite orig paper]]  & 1.2\\
     K$_{deep}$*\cite{ jimenez2018k} (augmented)[[cite orig paper]]  &  1.27\\
          Mid-level Fusion model$^{+}$\cite{jones2020improved} & 1.308 \\
     3D-CNN model$^{+}$\cite{jones2020improved} & 1.501 \\
     SG-CNN model$^{+}$\cite{jones2020improved} & 1.650 \\
     AK-score single\cite{kwon2020ak} & 1.415\\
     \hline
     AK-score ensemble\cite{kwon2020ak} & 1.293 \\ 
     
    \end{tabular}
    \caption{RMSE for test set using the standard protein binding training approach of train on the PDBBind refined dataset and test on the PDB core dataset and using with no validation. *score from the single best model [PDBBind v2016]}
    \label{tab:my_label}
\end{table}

\begin{table}[]
    \centering
    \begin{tabular}{|c|c|c|c|c|c|c|c|c|}
         train 1 & split 1 & shuffle & train 2 & split 2 & shuffle & gen 1 & gen 2 & epochs needed  \\
         \hline
         100\% & 80:10:10 & all & n/a & n/a & n/a& n/a & 0.915 & 29\\
         100\% & 80:10:10 & molecule & n/a & n/a & n/a & n/a &  & \\ 
        100\% & 80:10:10 & all & 100\% & 80:10:10 & molecule & n/a &  & \\ 
         75\% & 80:10:10 & all & 80:10:10 & 100\% & all & * &  0.904 & 64 (36+28)\\
        75\% & 80:10:10 & all & 80:10:10 & 100\% & molecule & * &  0.880 & 44 (32+12)\\
        50\% & 80:10:10 & all & 80:10:10 & 50\% & molecule & * &   & \\

    \end{tabular}
    \caption{The effect of training method and shuffling mode. *Needs to be gen tested.}
    \label{tab:training_method}
\end{table}

\clearpage

\section{Methods}

\begin{table}[]
    \centering
    \begin{tabular}{c|ccc}
        Network name & no. of layers & layer types & no of trainable parameters\\
        \hline
        sp-3 & 3 layers & sp-conv, average & \\
        deep two in net & 7 & sp-conv, fc, max pool, flatten, concat &\\
    \end{tabular}
    \caption{Layer count does not include flatten, concatiation or average pooling.}
    \label{tab:my_label}
\end{table}

\bibliographystyle{plain}
\bibliography{references}